\documentclass[sigconf]{acmart}

\renewcommand\footnotetextcopyrightpermission[1]{}

\AtBeginDocument{%
  \providecommand\BibTeX{{%
    \normalfont B\kern-0.5em{\scshape i\kern-0.25em b}\kern-0.8em\TeX}}}

\usepackage{graphicx}
\usepackage{latexsym}
\usepackage{adjustbox}
\usepackage{booktabs}
\usepackage{float}
\usepackage{url}
\usepackage{xspace}

\newcommand{\wv}{wav2vec2.0\xspace}
\newcommand{\mj}{Mockingjay\xspace}

\begin{document}

\title{What all do audio transformer models hear? Probing Acoustic Representations for language delivery and its structure}


\author{Jui Shah}
\authornotemark[1]
\email{jui.shah@midas.center}
\affiliation{%
  \institution{IIIT-Delhi}
}

\author{Yaman Kumar Singla}
\authornotemark[1]
\email{yamank@iiitd.ac.in}
\affiliation{%
  \institution{IIIT-Delhi, Adobe, State University of New York at Buffalo}
}
\author{Changyou Chen3}
\email{changyou@buffalo.edu}
\affiliation{%
  \institution{State University of New York at Buffalo}
}

\author{Rajiv Ratn Shah}
\email{rajivratn@iiitd.ac.in}
\affiliation{%
  \institution{IIIT-Delhi}
}


\begin{abstract}
  Transformer models across multiple domains such as natural language processing and speech form an unavoidable part of the tech stack of practitioners and researchers alike. Audio transformers that exploit representational learning to train on unlabeled speech have recently been used for tasks from speaker verification to discourse-coherence with much success. However, little is known about what these models learn and represent in the high-dimensional vectors. In this paper, we interpret two such recent state-of-the-art models, wav2vec2.0 and Mockingjay, on linguistic and acoustic features. We probe each of their layers to understand what it is learning and at the same time, we draw a distinction between the two models. By comparing their performance across a wide variety of settings including native, non-native, read and spontaneous speeches, we also show how much these modeles are able to learn transferable features. Our results show that the models are capable of significantly capturing a wide range of characteristics such as audio, fluency, suprasegmental pronunciation, and even syntactic and semantic text-based characteristics. For each category of characteristics, we identify a learning pattern for each framework and conclude which model and which layer of that model is better for a specific category of feature to choose for feature extraction for downstream tasks.
\end{abstract}

%% XXX



\maketitle
\pagestyle{plain} 
\section{Introduction}
Since the advent of transformers in the computational linguistics field in 2017 \cite{vaswani2017attention}, they have received great attention for a wide variety of tasks, ranging from constituency parsing \cite{kitaev2018constituency} to coherence modelling \cite{patil2020towards} and sentiment analysis \cite{tang2020dependency}. 
% outperforming the previous techniques. 
%\citet{tay2020efficient} survey prominent transformer models, which have now become a formidable force in the tech stack in  Natural Language Processing (NLP) \cite{devlinetal2019bert}, Computer Vision \cite{parmar2018image} and Reinforcement Learning \cite{parisotto2021efficient}. 
However, until recently the transformers have been limited to the discrete signal domain. Speech, being in the continuous domain, lags behind. 

As one of the first models for transformer based speech representation, vq-wav2vec \cite{baevski2019vq} proposed a two-stage pipeline. It discretizes an input speech to a $K$-way quantized embedding space (similar to word tokens for NLP tasks). The embeddings are then extracted from a BERT-based transfomer model. 
%However, this technique does not capture the context representation and dependencies across the time domain essential for continuous speech. 
Mockingjay \cite{liu2020Mockingjay} and AudioALBERT \cite{chi2020audio} are other such transformer models taking mel and fbank features as input, respectively. Mel-scale spectrogram as input are a more compendious acoustic feature compared to linear-scale spectrogram and fbank features are Mel filter bank coefficients which give better resolution at low frequencies and less at high frequencies, much like the human ear.
%These are modified versions of BERT for the audio domain. 
%They do not have an inbuilt feature extractor module. Hence, the former takes the input of 160-dim mel features and the latter takes in 160-dim fbank features. They share the same architecture with the difference being that AudioALBERT has shared parameters across the 12 encoder units, while for Mockingjay they are different. 
Wav2vec2.0 \cite{baevski2020wav2vec} is a recent transformer based speech representation model that converts an input audio to latent space embeddings via a contrastive task. %This task involves selecting the correct quantized latent representation of the masked time steps from a distractor set. 
 %which work three subunits - the feature encoder, the transformer, and the quantization module (discussed in Section \ref{sec:wav2vec2.0}). These units 

%Their inherent property to facilitate parallel training makes it easier to train models on large datasets. These pre-trained models are then fine-tuned on a variety of user-specific downstream tasks, achieving state-of-the-art results. 

These audio transformers have been applied over many diverse downstream speech language processing tasks with state-of-the-art results, such as speech translation \cite{wu2020self}, speaker recognition \cite{tian2020synchronous}, automatic scoring \cite{grover2020multi}, and sentiment classification \cite{tang2020dependency}. %is of utmost importance due to the unavailability of large amounts of labeled speech. 
This also begs the question as to what these transformer models are able to learn during the pretraining phase that helps them for various evaluation tasks\footnote{The sentiment of the above inquiry is also conveyed by Prof. Ray Mooney's quip that the meaning of a whole sentence cannot be captured by a \$\&!\#* vector \cite{conneau2018you,mooneyQuip}.}. %This question falls in the category of 
Besides, as more and more applications start relying on such models, it is important to explain what these embeddings capture to check for potential flaws and biases, which can affect a large number of applications. 

\begin{figure*}

  %\noindent
  \includegraphics[scale=0.75]{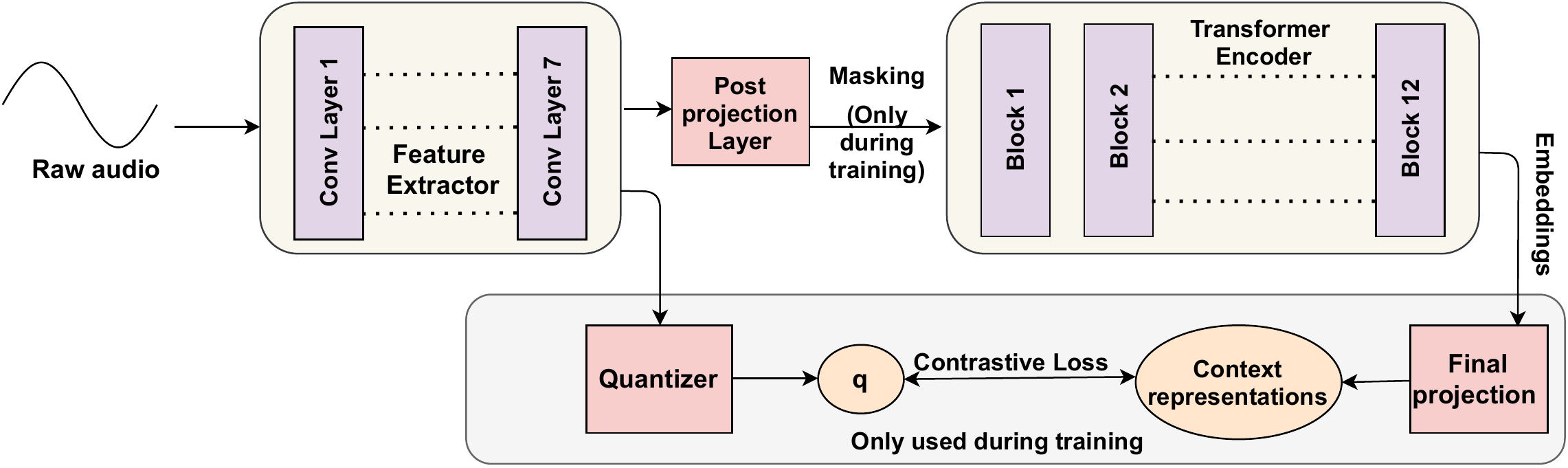}
  \caption{\label{fig:wv arch}wav2vec2.0  Architecture}
\end{figure*}

To this end, different research studies started probing language model embeddings for particular linguistic properties of interest. In \cite{belinkov2017neural}, Belinkov et al. probed for part-of-speech language understanding, \cite{hewitt2019structural} probed for syntax, \cite{peters2018deep} on morphology, \cite{zhang2020language} for scales and numbers, \emph{etc.} However, progress in the audio domain has been very limited with only a few works \cite{raj2019probing,alishahi2017encoding,belinkov2017analyzing,prasad2020accents}. Most of these works treat the audio encoders as automatic speech recognition (ASR) systems. Because of this restrictive treatment, they probe on a limited set of features important for ASR, such as phones, accent and style (spontaneous and non-spontaneous). However, the analysis does not explain the state-of-the-art performance that audio encoders achieve on a wide variety of tasks.

% \cy{Please rewrite: On the other hand, transformer models are still fairly recent in the audio domain primarily self supervised learning in NLP and Vision belongs to the discrete domain whereas speech spans in the continuous domain}.
%Transformers have predominantly addressed the discrete data domain. Hence, NLP and vision fields oversaw a tremendous amount of work on transformer-based modelling. 

% model-interpretability which has gained new found focus recently \cite{doshi2017towards}.

%Several attempts to interpret BERT (Bidirectional Encoder Representations from Transformers) \cite{devlinetal2019bert}, which has become the \textit{de-facto} model for most of NLP models, have been made such as in \cite{jawahar-etal-2019-bert,cui2020does,ramnath2020towards}. \citet{jawahar-etal-2019-bert} probe each of the different layers of BERT to find which layers best learn the phrase-level information, linguistic information and the long-distance dependencies. The results showed what role each layer played and the study concluded that the middle layers learnt the syntactic features and the higher levels learnt the semantic features and that the deeper layers are needed for long-distance dependencies while the initial layers capture the phrase-level information. Meanwhile, no probe has been done for representation learning Audio transformers.

Our contributions are summarized as: 
(1)~We introduce here (47) probing tasks to capture simple linguistic features of speech audios, and we use them to study embeddings generated by two different audio transformers on three types of speeches, uncovering intriguing properties of encoders.

(2)~We propose a detailed analysis of what is learned by the recent transformer-based semisupervised audio encoder models, {\wv} and {\mj}. We implement post hoc probing on the embeddings extracted from each intermediate unit of the two models. We probe these embeddings using an extensive diversity (4 high-level categories) and number of features (46 in total), each categorized by the linguistic property they probe. 
%We do this for text-based, audio-based, vocabulary-based, fluency-based, and suprasegmental pronunciation-based features. 
We extract the results on all the features relevant to speech covering both \textit{what} was spoken and \textit{how} it was spoken. These results help us lay out a map of what particular features are learned in each layer while also providing a metric of comparison between the two models. These features are crucial for downstream applications such as automatic scoring, readability evaluation, automatic speech generation quality, text to speech quality, accent detection, ASR models, \textit{etc.} \cite{yan2018complexity,barth2014effects,rasinski2004assessing,zhang2019learning,kyriakopoulos2020automatic,jyothi2015improved}. As a proof of concept, we also show the effect of our analysis on two such downstream applications (speaker identification and phone classification) (\S\ref{sec:Effect on Downstream Tasks}).

(2)~We test the models for their representative effectiveness on different types of speech settings: native-read, native-spontaneous, and non-native-read. We find that, for the most part, native-spontaneous and non-native speech settings follow the result patterns for native-read dataset albeit with a worse performance. In general, type of speakers matter less than the type of speech. %Therefore, for both the models, we observe that non-native read speech performs better than spontaneous speech in general.

(3)~We identify the role of the feature extractor module in wav2vec2.0, which enables it to process raw input audio of $16 KHz$ without any preprocessing. We find that the subsequent layers of the feature encoder can encode all features into increasingly dense and informative representation vectors without any ``intelligent processing'' on them.

(4)~We compare the performance of the representations from audio models and BERT on text features. %\cy{how can you compare with BERT as BERT is for text data, and your model is for audio data, I believe.}. 
This is the first work to check the representative capacity of audio representations for the text captured by audio. We find that despite of having no text-specific error metrics, the audio models are able to encode text well and are comparable to BERT on several parameters. We find that the dataset used to pre-train audio models has a significant effect on downstream performance. %Surprisingly, while both {\wv} and {\mj} outperform BERT on LibriSpeech (the dataset they were trained on), they underperform in other settings. Additionally, both models seem to learn surface-level text features (such as the number of nouns and pronouns) comparable to BERT.

% in the hope that, similar to the extensive work done on text transformer models, audio models in the speech community are also developed and interpreted. 
To the best of our knowledge, this is the first attempt towards interpreting audio transformer models\footnote{We will release our code, datasets and tools used to perform the experiments and inferences upon acceptance.}. The conclusion points out that the transformers are able to learn a holistic range of features, which enable them to perform with great accuracy on various downstream tasks even training solely on unlabeled speech. 
%%%%%%%%%%%%%%%%%%%%%%%%%%%%%%%%%%%%%%%%%%%%%%%%%%%%%%%%%%%%%%%%%%%%%%%%%%%
%%%%%%%%%%%%%%%%%%%%%%%%%%%%%%%%%%%%%%%%%%%%%%%%%%%%%%%%%%%%%%%%%%%%%%%%%%%

\section{Brief Overview Of The Probed Models}
\label{sec:Brief Overview Of The Probed Models}
We probe three recent transformer based models: \wv, \mj and BERT. Below, we give a brief overview of the three models and their high-level architectures.
%%%%%%%%%%%%%%%%%%%%%%%%%%%%%%%%%%%%%%%%%%%%%%%%%%%%%%%%%%%%%%%%%%%%%%%%%%%
%%%%%%%%%%%%%%%%%%%%%%%%%%%%%%%%%%%%%%%%%%%%%%%%%%%%%%%%%%%%%%%%%%%%%%%%%%%
\subsection{wav2vec2.0}
\label{sec:wav2vec2.0}

\begin{figure*}
  %\noindent
  \includegraphics[scale=0.75]{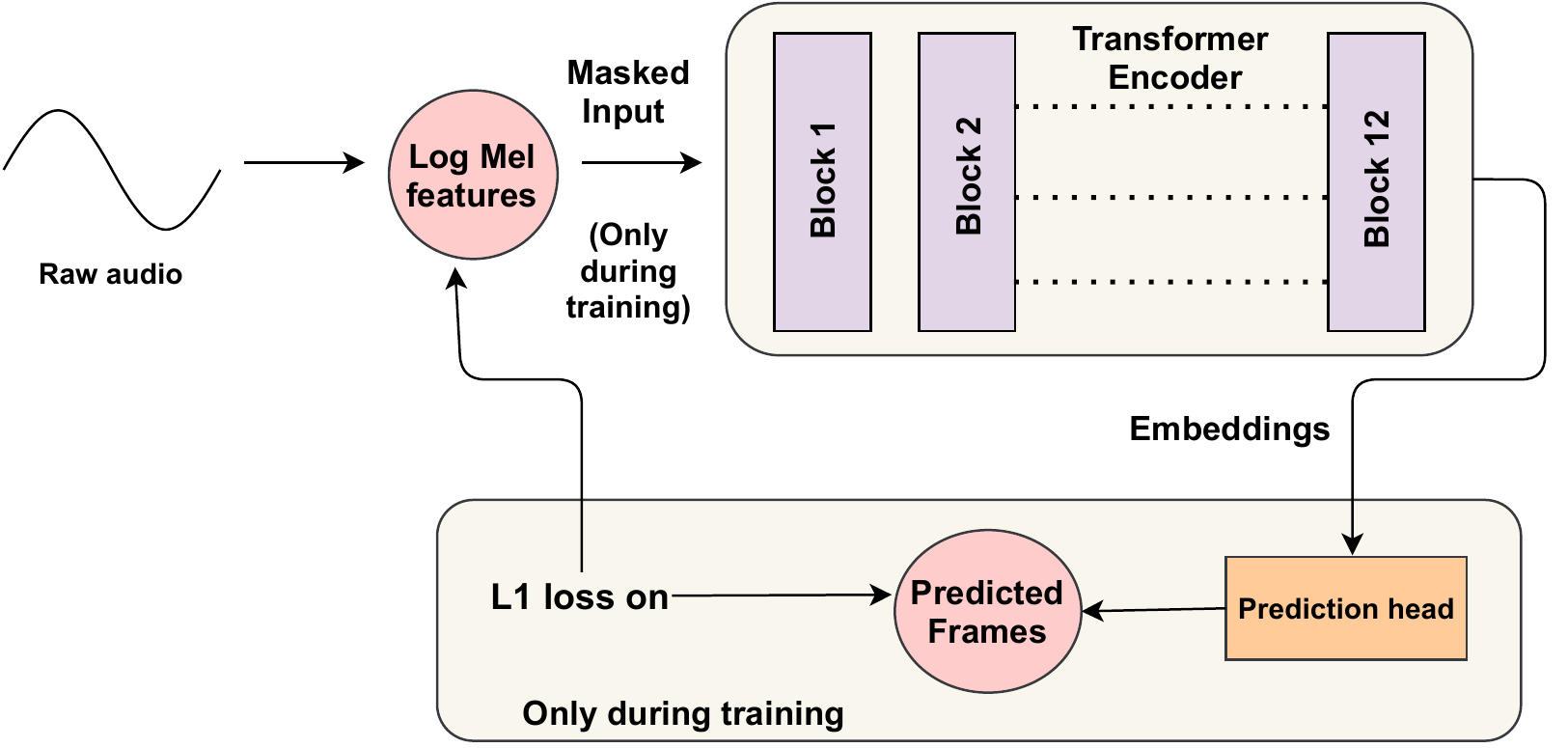}
  \caption{Mockingjay Architecture}
\end{figure*}

{\wv} is a recent transformer based speech encoding model. It is composed of 3 major components - the feature encoder, the transformer, and the quantization module. The feature encoder consists of a multi-layer convolutional network which converts the raw input audio input $X$ to latent representation ${Z_{1}, Z_{2},..,Z_{t}}$. These latent vectors are fed into the transformer to build the representations $C_{1},C_{2},...C_{n}$. The training is done by masking certain time-steps in the latent feature representation and learning a contrastive task over it. The contrastive task requires finding the correct quantized representation corresponding to the masked latent audio representation amongst a set of distractors. The contrastive task targets ($q_t$) are built by passing the output of feature encoder to the quantizater at various time steps.

The model is pretrained on unlabeled Librispeech data \cite{panayotov2015librispeech} and then finetuned on TIMIT \cite{garofolo1993darpa} dataset for phoneme recognition. It achieves a 1.8/3.3 WER on the clean/noisy test sets on experiments using all labeled data of Librispeech and 5.2/8.6 WER on the noisy/clean test sets of Librispeech using just ten minutes of labeled data. The authors claim that even while lowering the amount of labeled data to one hour, \wv outperforms the previous state of the art on the 100 hour subset while using 100 times less labeled data.

All our experiments are based on the \wv-base model in which the feature encoder contains $7$ blocks having a temporal convolution of 512 channels with strides $(5,2,2,2,2,2,2)$ and kernel widths $(10,3,3,3,3,2,2)$ respectively and the there are $12$ transformer blocks with a model dimension of $768$, inner dimension (FFN) $3,072$ and $8$ attention heads.

A point to note is that the output of each of the transformer block depends on the duration of the audio file. For a moderate size audio ($\sim$5 seconds), the embedding obtained is huge in size. It is of the form $768 * T$ where T is dependent on the duration of the audio. Hence, to probe the different features, we time-average the embeddings.

%%%%%%%%%%%%%%%%%%%%%%%%%%%%%%%%%%%%%%%%%%%%%%%%%%%%%%%%%%%%%%%%%%%%%%%%%%%
%%%%%%%%%%%%%%%%%%%%%%%%%%%%%%%%%%%%%%%%%%%%%%%%%%%%%%%%%%%%%%%%%%%%%%%%%%%

\subsection{Mockingjay}
\label{sec:mockingjay}
{\mj} is a bidirectional transformer model which allows representation learning by joint conditioning on past and future frames. It accepts input as 160 dimension log-Mel spectral features\footnote{For a primer on log-Mel and other audio feature extraction, refer to \cite{choi2017tutorial}} and has outperformed it for phoneme classification, speaker recognition and sentiment discrimination accuracy on a spoken content dataset by  35.2\%, 28.0\% and 6.4\% respectively. The authors claim the model is capable of improving supervised training in real world scenarios with low resource transcribed speech by presenting that the model outperformes other exisiting methods while training on 0.1\% of transcribed speech as opposed to their 100\%.

For our experiments, we use the MelBase-libri model. The architecture comprises of $12$ encoder layers and each unit has 0he same output dimension of $768$ and comprises of sub-layers which include a feed-forward layer of size $3072$ and $12$ self-attention heads.  We probe each of the $12$ transformer blocks of both models and the feature encoder of wav2vec2.0 to check if they learn the features of audio, fluency, suprasegmental pronunciation and text.

Similar to {\wv}, {\mj} also has huge embeddings of size $768*T$ with T dependent on the size of audio.
%%%%%%%%%%%%%%%%%%%%%%%%%%%%%%%%%%%%%%%%%%%%%%%%%%%%%%%%%%%%%%%%%%%%%%%%%%%
%%%%%%%%%%%%%%%%%%%%%%%%%%%%%%%%%%%%%%%%%%%%%%%%%%%%%%%%%%%%%%%%%%%%%%%%%%%
\subsection{BERT}
\label{sec:BERT}
BERT stands for Bidirectional Encoder Representations and proved to be a major breakthrough for NLP. The architecture basically comprises encoder layers stacked upon eachother. BERT-Base has $12$ such layers while BERT-Large has $24$. We have probed the uncased Base model. The input format to the transformer has $3$ parts - a classification token(CLS), sequence of words and a separate sentence(SEP) token. The feed-froward network has $768$ hidden units and $12$ attention heads. BERT achieves effective performance on various NLP tasks. Similar to audio models, we probe BERT extracting embeddings from each of the $12$ encoder blocks. Since, text has no time component, the embeddings are of size $768*1$.

%%%%%%%%%%%%%%%%%%%%%%%%%%%%%%%%%%%%%%%%%%%%%%%%%%%%%%%%%%%%%%%%%%%%%%%%%%%
%%%%%%%%%%%%%%%%%%%%%%%%%%%%%%%%%%%%%%%%%%%%%%%%%%%%%%%%%%%%%%%%%%%%%%%%%%%

\section{Probing - Problem Definition And Setup}
\label{sec:Probing - Problem Definition And Setup}
Here we specify the probing model and explain how we compare the audio and text transformer models. We also give an overview of all the features and models we probe in the paper along with the datasets used.
    
\begin{figure*}
  %\noindent
  \includegraphics[scale=0.6]{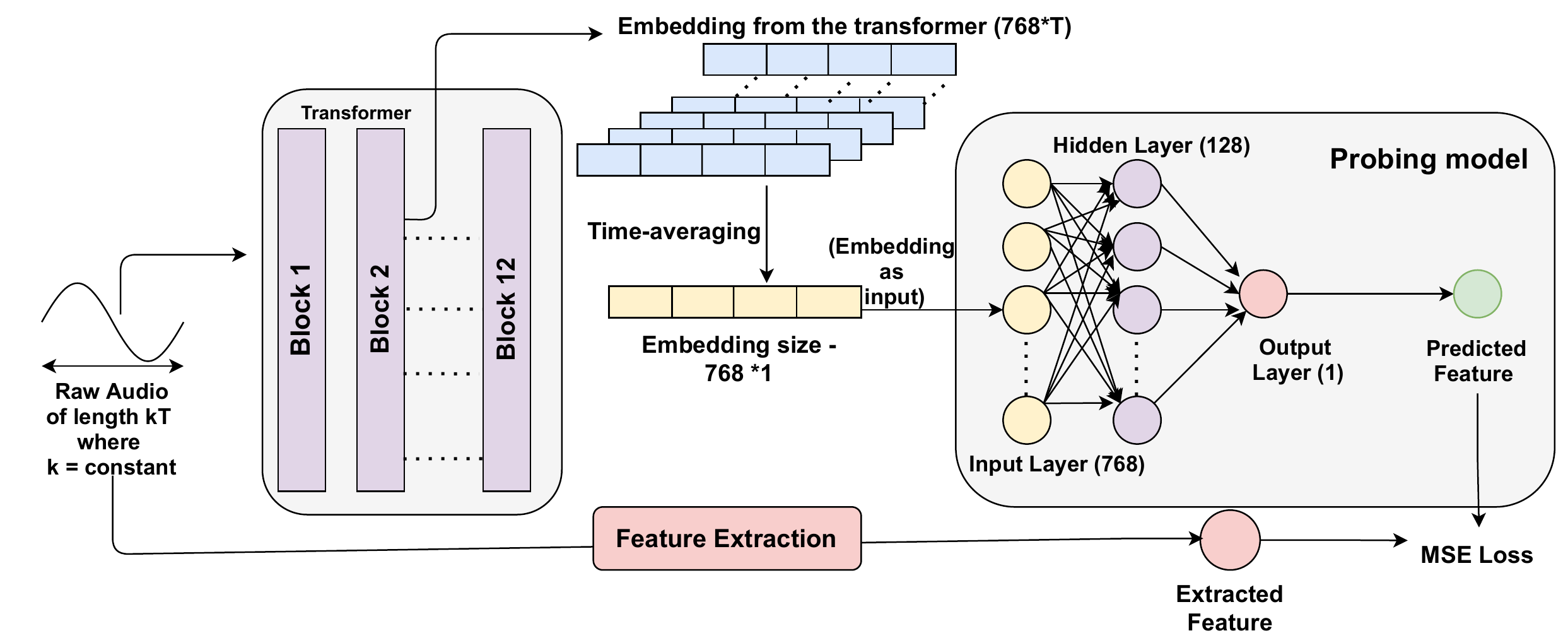}
  \caption{\label{fig:Procedure for probing audio transformers}Procedure for probing audio transformers}
\end{figure*}

\subsection{Probing Model}
\label{sec:Probing Model}
We define the problem of probing a model $M$ for a feature $f$ as a regression task using a probing model $P$. $P$ is a 3-layer feed forward neural network trained on $M$'s emebddings to predict the feature $f$. For instance, in text-transformers, a probing model ($P$) might map BERT embeddings ($M$) to syntactic features such as parts of speech ($f$) \cite{jawahar-etal-2019-bert}. Post model training, the representational capacity of embeddings is judged based on the ease with which the 3-layer feed-forward probe network is able to learn the said feature. Metrics like accuracy and MSE loss are used for measuring and comparing the representational capacities \cite{alishahi2017encoding,belinkov2017analyzing,prasad2020accents,belinkov2017neural,jawahar-etal-2019-bert}.

Our probe model consists of a $3$-layer fully connected neural network with the hidden layer having a ReLU activation and dropout to avoid over-fitting\footnote{Model dimensions are $(768,128,1)$ for all the intermediate layers of Transformers and $(512,128,1)$ for the feature extractor. Adam with a learning rate of $0.0001$ is used.}. We compare the representative capacity of different audio and text transformers on the basis of the loss values reported by the prober. Furthermore, we take a randomly initialized vector as a baseline to compare against all the `intelligent' models. This approach is in line with some of the previous works in the model interpretability domain \cite{alishahi2017encoding,belinkov2017analyzing,prasad2020accents,belinkov2017neural,jawahar-etal-2019-bert}. A diagram explaining the overall process is given in the Figure~\ref{fig:Procedure for probing audio transformers}.

\subsection{Feature Overview}
\label{sec:Feature Overview}
We test the audio transformer models on the following speech features: audio features (\S\ref{subsec:Audio Features}), fluency features (\S\ref{subsec:Fluency Features}), and pronunciation features (\S\ref{subsec:Pronunciation Features}). Since spoken language can be considered as a combination of words (\textit{what} was spoken), and language delivery (\textit{how} it was spoken), we probe audio transformer models for both speech and text knowledge. For comparing on textual representational capacity, we extract text features from the original transcripts of all the audio datasets considered (\S\ref{sec:Can Audio Models Read Too?}).  %probing on text features extracted from Wikipedia articles (randomly chosen $2000$ sentences for training and $200$ for testing).\footnote{For these experiments, we convert Wikipedia articles to speech using the Google text-to-speech model \cite{gtts} and use that to feed the generated audio to the audio transformer models.} 
A detailed description of all features extracted and their methodology of extraction is given in Section~\ref{sec:What Do Audio Transformers Hear?} (audio features) and Section~\ref{sec:Can Audio Models Read Too?} (text features).

\subsection{Types of Speech Explored} 
\label{sec:Types of Speech Explored}
Unlike text, speech varies drastically across speech types. For instance, a model developed for American (native) English speakers produces unintelligible results for Chinese (non-native) English speakers \cite{mulholland2016comparison}. Since transformer models tend to be used across multiple speech types \cite{grover2020multi,doumbouya2021using}, it is important to assess and compare their performance and bias across each of the speech types. Therefore, we test them on native read, native spontaneous, and non-native read speech corpora.

For probing on native read speech, we use the LibriSpeech dataset \cite{panayotov2015librispeech}. We take the default `train-clean-100' set from LibriSpeech for training the probing model and the `test-clean' set for testing it. For native spontaneous English speech, we use the Mozilla Common Voice dataset \cite{ardila2019common}. We use a subset of $2000$ random audios for training and $200$ audios for testing.
For interpreting audio transformers on non-native speech, we use L2-Arctic dataset \cite{zhao2018l2arctic}. We take $500$ audios of $4$ speakers each for training the prober and $50$ audios each for testing. The $4$ speakers are selected in such a way that there is $1$ male and $1$ female speaker each with Hindi and Spanish as their first languages.

\subsection{Models Probed}
\label{sec:models probed}
We probe two recent audio transformers, \wv and \mj for their speech and language representational capacities. For text-based linguistic features particularly, we also compare them with BERT embeddings \cite{devlinetal2019bert}. See Section~\ref{sec:Brief Overview Of The Probed Models} for an overview of the three transformer models. 

Self-attention is the powerhouse which drives these transformers \cite{vaswani2017attention}. It is the main reason behind their state-of-the-art performance on diverse tasks. While \mj is exclusively built of self-attention and feed-forward layers, \wv also has several CNN layers. They are presented as ``feature extractor'' layers in the original paper (Figure~\ref{fig:wv arch}). Therefore, we also investigate the role of the feature extractor in wav2vec2.0. In particular, we investigate that whether similar to computer vision \cite{erhan2010understanding,krizhevsky2012imagenet,girshick2014rich}, do the CNN layers in speech transformers also learn low-level to high-level features in the subsequent layers. Very few studies in the speech domain have tried to answer this question \cite{zhang2018visualization}. %We also include the post projection layer - the layer after the feature encoder but before the transformer, which maps the dimensions of extracted features $(512)$ to that of the Transformer $(768)$.

%Similar to previous works on model interpretability of text transformers \cite{jawahar-etal-2019-bert}, we evaluate the performance of deep representations on the mean-square error for each encoder unit. 

We probe the representational capacity of embeddings from all layers of the three transformer models. This helps us understand the transformer models at four levels, \textit{i.e.}, across models, speech types, input representations (log Mel and raw audio), and layers. This analysis gives us results on a much finer level than just comparing the word error rates of the two models. It helps us to know the linguistic strengths and weaknesses of the models and how they are structuring and extracting information from audio. We also use our interpretability results to improve the performance on some downstream tasks (\S\ref{sec:Effect on Downstream Tasks}).

%%%%%%%%%%%%%%%%%%%%%%%%%%%%%%%%%%%%%%%%%%%%%%%%%%%%%%%%%%%%%%%%%%%%%%%%%%%
%%%%%%%%%%%%%%%%%%%%%%%%%%%%%%%%%%%%%%%%%%%%%%%%%%%%%%%%%%%%%%%%%%%%%%%%%%%

\begin{figure*}[htbp]
\centering
    %\vspace{-2mm}
    \includegraphics[scale=0.5]{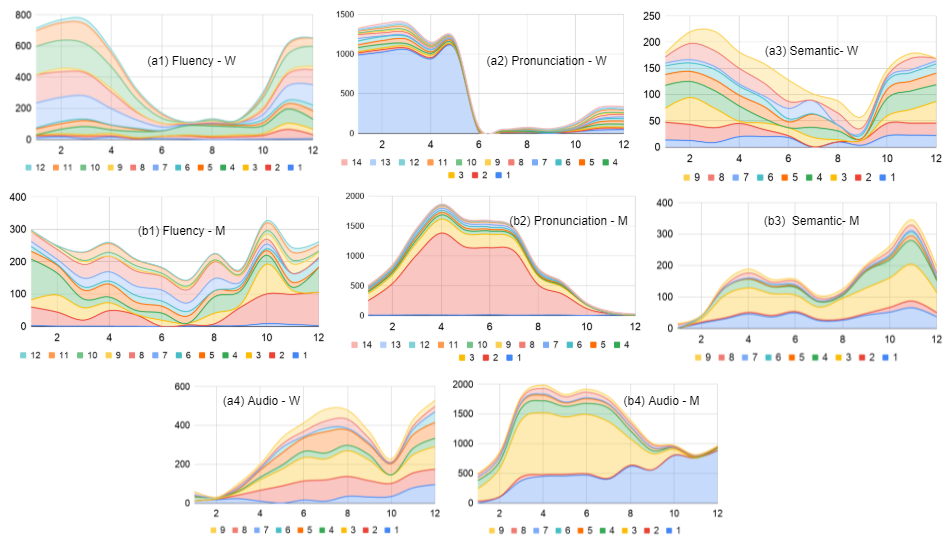}
    %\vspace{-8mm}
    \caption{
    \label{figure:Comparison of w2v and MJ on native read speech}
    \small
    Performance of \wv and \mj on fluency features (a1,b1), pronunciation features (a2,b2), surface level text features (a3,b3), audio features (a4,b4) (`W' denotes \wv and `M' denotes \mj). The graphs represent stacked area charts with the x-axis being the layers of the model and y-axis shows the relative performance of each layer with respect to the maximum loss for each feature \emph{((loss - min\_loss)*100\%/min\_loss)}. Hence, higher the value, higher the loss, lower the performance. %Different colours symbolize the curves followed by the different features across the transformers.
    \newline The feature numbers according to category are given below:
    \newline
    \textbf{Audio features:}~ \emph{1. total duration, 2. stdev energy, 3. mean pitch, 4. voiced to unvoiced ratio, 5. zero crossing rate, 6. energy entropy, 7. spectral centroid, 8. localJitter, 9. localShimmer},
    \newline
    \textbf{Fluency features:}~\emph{1. filled pause rate, 2. general silence, 3. mean silence, 4. silence abs deviation, 5. SilenceRate1, 6. SilenceRate2, 7. speaking rate, 8. articulation rate, 9. longpfreq, 10. average syllables in words, 11. wordsyll2, 12. repetition freq},
    \newline
    \textbf{Pronunciation features:}~\emph{1. StressedSyllPercent, 2. StressDistanceSyllMean, 3. StressDistanceMean, 4. vowelPercentage, 5. consonantPercentage, 6. vowelDurationSD, 7. consonantDurationSD, 8. syllableDurationSD, 9. vowelSDNorm, 10. consonantSDNorm, 11. syllableSDNorm, 12. vowelPVINorm, 13. consonantPVINorm, 14. syllablePVINorm}, and 
    \newline
    \textbf{Semantic level text features:}~\emph{1. Total adjectives, 2. Total adverbs, 3. Total nouns, 4. Total verbs, 5. Total pronoun, 6. Total conjunction, 7. Total determiners, 8. Number of subjects, 9. Number of objects}
}
%\vspace{-5mm}
\end{figure*}

%%%%%%%%%%%%%%%%%%%%%%%%%%%%%%%%%%%%%%%%%%%%%%%%%%%%%%%%%%%%%%%%%%%%%%%%%%%
%%%%%%%%%%%%%%%%%%%%%%%%%%%%%%%%%%%%%%%%%%%%%%%%%%%%%%%%%%%%%%%%%%%%%%%%%%%

\section{What Do Audio Transformers Hear?}
\label{sec:What Do Audio Transformers Hear?}

%The audio transformers {\mj} and {\wv} have enabled speech recognition models at low word error rate by training on just a few minutes of transcribed speech and hours of unlabeled speech. In order to understand what these models learn with much less annotated data via their self-supervision task, we assess the models on knowledge of language delivery and content features. Hence, we assesses both \emph{what} was spoken and \emph{how} it was spoken.

In this section, we probe audio (\S\ref{subsec:Audio Features}), fluency (\S\ref{subsec:Fluency Features}), and pronunciation (\S\ref{subsec:Pronunciation Features}) features. These features are extracted directly from the audio waveform. Amongst them, the audio features measure the knowledge of the core features of audio including energy, jitter, shimmer and duration. Fluency features measure the smoothness, rate, and effort required in speech production \cite{de2009praat,yan2018complexity}. Pronunciation features measure the intelligibility, accentedness and stress features of the audio. Tasks such as automatic scoring, readability evaluation, automatic speech generation quality, text to speech quality, accent detection, ASR models, \textit{etc.} are impacted by the fluency and pronunciation features \cite{yan2018complexity,barth2014effects,rasinski2004assessing,zhang2019learning,lee2019robust,kyriakopoulos2020automatic,jyothi2015improved}. 

A typical embedding of the transformers at any layer is of the size $768*T$ where T depends on the duration of the speech segment. We average it to get $768*1$ dimension embedding which serves as the representation of the speech segment for which we have extracted the features. This is then fed as the input to our probing model. Figure~\ref{fig:Procedure for probing audio transformers} depicts the process.

%By comparing losses, we identify learning patterns of the two representational-learning based audio transformer models - {\wv} and {\mj}. We also draw out a comparison between the two. Furthermore, we compare the performances of the same models across various types of dataset - native read, native spontaneous and non-native read speech to detect bias.
%Additionally, we also study the role of the Feature Encoder(FE) module in {\wv} including the post projection layer of {\wv} (transforms the $512$ dimension to the encoder input dim of $768$. The feature) in the module in the FE module itself. FE is especially important since it converts the raw audio given to {\wv} to a vector representation in the model itself unlike {\mj} which uses Mel features as input directly.

%%%%%%%%%%%%%%%%%%%%%%%%%%%%%%%%%%%%%%%%%%%%%%%%%%%%%%%%%%%%%%%%%%%%%%%%%%%
%%%%%%%%%%%%%%%%%%%%%%%%%%%%%%%%%%%%%%%%%%%%%%%%%%%%%%%%%%%%%%%%%%%%%%%%%%%

\subsection{Audio knowledge}
\label{subsec:Audio Features}

\begin{table*}[]
\begin{tabular}{@{}|l|l|l|@{}}  \toprule
\textbf{Audio feature} & \textbf{Description} & \textbf{Extracted Using}\\ \midrule \midrule
Total duration & Duration of audio &  Librosa \cite{mcfee2015librosa}\\ \midrule
zero-crossing rate & Rate of sign changes & PyAudioAnalysis \cite{giannakopoulos2015pyaudioanalysis}\\ \midrule
energy entropy & Entropy of sub-frame normalized energies & PyAudioAnalysis \cite{giannakopoulos2015pyaudioanalysis} \\\midrule
spectral centroid & Center of gravity of spectrum & PyAudioAnalysis \cite{giannakopoulos2015pyaudioanalysis} \\\midrule
mean pitch & Mean of the pitch of the audio &  Parselmouth \cite{parselmouth,praat}\\ \midrule
local jitter & Avg. absolute difference between consecutive\\  & periods divided by the avg period & Parselmouth \cite{parselmouth,praat}\\ \midrule
local shimmer & Avg absolute derence been\\ & the amplitudes of consecutive periods, & Parselmouth \cite{parselmouth,praat}\\ & divided by the average amplitude &  \\ \midrule
voiced to unvoiced ratio & Number of voiced frames upon\\ & number of unvoiced frames & Parselmouth \cite{parselmouth,praat}\\ \bottomrule
\end{tabular}
\caption{ \label{table: audio feature extraction}Audio feature extraction algorithms and libraries used}
\end{table*}

We measure the following audio features: \emph{Total duration, zero-crossing rate, energy entropy, spectral centroid, mean pitch, local jitter, local shimmer, and voiced to unvoiced ratio}. \emph{Total duration} is a characteristic feature of the audio length that tells us about the temporal shape of the audio. The temporal feature \emph{zero crossing rate} measures the rate at which a signal moves from positive to a negative value or vice-versa. It is widely used as a key feature in speech recognition and music information retrieval \cite{neumayer2007integration,simonetta2019multimodal}. Energy features of audio are an important component that characterizes audio signals. We use \emph{energy entropy} and the standard deviation of energy (\emph{std\_dev energy}) to evaluate the energy profile of audio. \emph{Spectral centroid} is used to characterise the spectrum by its centre of mass. To estimate the quality of speech as perceived by the ear, we measure the \emph{mean pitch}. We also probe for frequency instability (\emph{localJitter}), amplitude instability (\emph{localShimmer}), and \emph{voiced to unvoiced ratio}. Table~\ref{table: audio feature extraction} mentions the libraries and algorithms used for extracting the above features. Next we present the results of our probing experiments on the two transformers for three different speech types.

\textbf{Native Read Speech:} 
Figures~\ref{figure:Comparison of w2v and MJ on native read speech}(a4,b4)\footnote{Refer Tables~\ref{Audio_W} and \ref{Audio_M} of Appendix for loss values} shows the results obtained for audio features probed on wav2vec2.0 and Mockingjay on the Librispeech dataset. It can be seen that the lowest loss is obtained in the initial two layers for wav2vec2.0, whereas it is the final layer for Mockingjay. These results also indicate that unlike computer vision there is no uniform conception of ``high-level'' or ``low-level'' in audio transformers \cite{erhan2010understanding,krizhevsky2012imagenet,girshick2014rich}. We can see a clear ascent in the losses as we traverse the graph for wav2vec2.0 from left to right, \emph{i.e.}, from lower layers to the higher layers. This suggests that as we go deeper into the $12$ block transformer model the audio features are diluted by \wv. \mj, on the other hand, follows a negative slope for its losses from the first to the last layers. Hence, the audio features are best captured in the final layers of the Mockingjay model. 

When comparing the minimum losses across both models, the average learning of these features for \wv is better than that of \mj by $28.59\%$. Even with the final layer embedding, wav2vec2.0 performs better than Mockingjay by $24.53\%$. This is interesting given that the final layer of wav2vec2.0 contains the most diluted version of the learned features and Mockingjay has its best version (in the final layers). Therefore, \wv has richer audio representations compared to \mj.

\begin{figure*}
  %\noindent
  \includegraphics[scale=0.5]{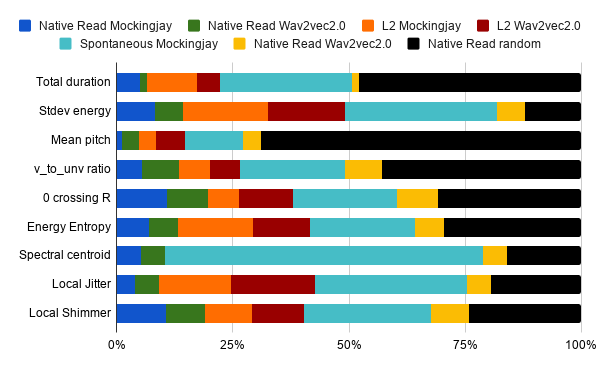}
  \caption{Performance of each audio feature (on the y-axis) relative to the performance of random embeddings on the three speech types (native read, native spontaneous, and non native speech). X-axis represent the MSE loss values relative to random embeddings loss \emph{(loss*100/l2\_random\_loss)}.}
  \label{figure: w2v and mj on audio features compared with random embeddings}
\end{figure*}

\textbf{Native Spontaneous Speech:} 
For native spontaneous speech, as shown in Figure~\ref{figure: w2v and mj on audio features compared with random embeddings}\footnote{Refer Tables~\ref{nsAudio_W} and \ref{nsAudio_M} of Appendix for loss values}, \wv is observed to perform better than \mj. Wav2vec2.0, on an average performs better by $41.69\%$ when compared across the best performing layers and $51.12\%$ when end layer losses are compared. The pattern of the best performing layer also remains the same as the case of native read speech for \mj. %Mockingjay learns these features best in the second last (11\textsuperscript{th} layer) for native spontaneous speech, while it was the last two layers for native read speech. 
For wav2vec2.0, native read speech was best captured in the initial $2$ layers, but for spontaneous speech, the layers are a bit more spread out across the initial half of the transformer model. We also observe that the loss values on native spontaneous speech are higher than the ones for native read and non-native read corpora.

\textbf{Non-native Speech:} When tested on L2 speakers (Figure~\ref{figure: w2v and mj on audio features compared with random embeddings}\footnote{Refer Tables~\ref{nAudio_W},~\ref{nAudio_M} of Appendix for loss values}), wav2vec2.0 outperforms Mockingjay by $9.53\%$ and $12.51\%$ on minimum and end layer losses, respectively. Additionally, similar to the case of native read speech, Mockingjay learns the audio features best in the final layers. As for wav2vec2.0, the layers learning the audio features are spread out with the initial half of the model learning them more accurately than the later half.

%%%%%%%%%%%%%%%%%%%%%%%%%%%%%%%%%%%%%%%%%%%%%%%%%%%%%%%%%%%%%%%%%%%%%%%%%%%
%%%%%%%%%%%%%%%%%%%%%%%%%%%%%%%%%%%%%%%%%%%%%%%%%%%%%%%%%%%%%%%%%%%%%%%%%%%

\subsection{Fluency knowledge}
\label{subsec:Fluency Features}
To the best of our knowledge, we use the features that measure fluency for the first time in this paper. 
The key features of fluency are: rate of speech, pauses, and length of runs between pauses \cite{yan2018complexity}. To measure the rate of speech, we measure the speech rate (number of words per second in the total response duration) (\emph{speaking\_rate}) and articulation rate (number of words per second in the total articulation time, \emph{i.e.,} the resulting duration after subtracting the time of silences and filled pauses from the total response duration) (\emph{articulation\_rate}) \cite{wood2001search}. Apart from these rates, pauses in speech are the second most observable feature to indicate disfluency \cite{igras2016structure}. Therefore, we measure the duration, location and frequency of pauses as prototypical features. For this, we measure the number of filled pauses per second - (\emph{filled\_pause\_rate}), \emph{silence deviation} (absolute difference from the mean of silence durations), which along with the total duration of the audio helps to indicate the length of runs between the pauses \cite{mohle1984comparison}. This also serves an important indicator for fluency. Other features include total number of silences (\emph{general silence}), mean duration of silences (\emph{mean\_silence}), average silence per word (\emph{SilenceRate1}), average silence per second (\emph{SilenceRate2}) and number of long silence per word (\emph{longpfreq}).

\begin{table*}[]
\begin{tabular}{@{}|l|l|@{}}  \toprule
\textbf{Fluency feature} & \textbf{Description}\\ \midrule \midrule
Filled pause rate & Number of filled pauses (uh, um) per second 
\cite{parselmouth,praat}\\ \midrule
General silence & Number of silences where silent duration between two words \\ & is greater than 0.145 seconds \\ \midrule
Mean silence & Mean duration of silence in seconds \\ \midrule
Silence abs deviation & Mean absolute difference of silence durations \\ \midrule
Silence rate 1 & Number of silences divided by total number of words\\ \midrule
Silence rate 2 & Number of silences divided by total response duration in seconds \\ \midrule
Speaking rate & Number of words per second in total response duration\\ \midrule
Articulation rate & Number of words per second in total articulation time 
                                   (i.e. the resulting \\ & length of subtracting the time of silences and filled pauses 
                                   from the \\ & total response duration).\\ \midrule
Long pfreq & Number of long silences per word\\ \midrule
Avg syllables in words & Get average count of syllables in words after removing
                                                    all stop words\\ & and pause words.\\ \midrule
Word syll2 & Number of words with syllables greater than two\\ \midrule
Repetition freq & Frequency of repetition by calculating number of repetition\\ & divided by total number of words.\\
\bottomrule
\end{tabular}
\caption{ \label{table: fluency feature extraction}Fluency feature extraction algorithms and libraries used for extracting them are numpy, textgrids.
}
\end{table*}

Furthermore, conversational fillers are a major source of disfluency. Sounds like \emph{uh, um, okay, you know, etc} are used to bring naturalness and fluency to their speech. The extent of fillers is an important feature to check for speech fluency. We use the average number of syllables in a word (\emph{average\_syllables\_in\_word}), the number of words with syllables greater than 2 (\emph{wordsyll2}) and the repetition frequency (\emph{repetition\_freq}), to measure this.

\begin{figure*}
  %\noindent
  \includegraphics[scale=0.5]{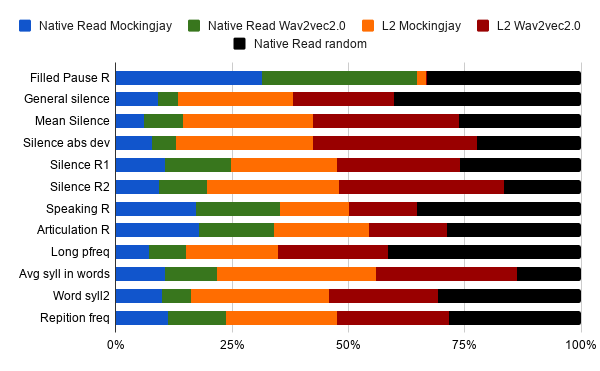}
  \caption{\label{Fluency}Performance of each fluency feature (on the y-axis) relative to the the performance of random embeddings on L2 Arctic data features \emph{(loss*100/l2\_random\_loss)} on the x-axis where loss values are that of MSE}
\end{figure*}

\textbf{Native Read Speech:}
For fluency based features on native read speech, similar to audio features, wav2vec2.0 performs better than Mockingjay (Figures~\ref{figure:Comparison of w2v and MJ on native read speech} (a1) and (b1)\footnote{Refer Tables \ref{Fluency_W} and \ref{Fluency_M} of Appendix for loss values}). While the fluency features are not layer specific but are spread across the model for Mockingjay, they tend to show the best performance in the middle layers for wav2vec2.0. With the final layer embeddings of both models, wav2vec2.0 performs better than Mockingjay by $12.23\%$. The performance gap increases by four folds to $42.37\%$ when compared on the minimum losses (among all observed for the intermediate layers) learnt by both models.

\textbf{Non-native Speech:} For the L2 Arctic dataset (\footnote{Refer Tables \ref{nFluency_W} and \ref{nFluency_M} of Appendix for loss values}), the learning of fluency features is concentrated in the middle layers for wav2vec2.0. Moreover, here we see a definite pattern that Mockingjay is learning better in the final layers compared to the no pattern observed in the case of Librispeech. Overall, wav2vec2.0 outperforms Mockingjay by $5.06\%$ on the minimum loss layers but by $105.62\%$ for the final layers. Thus, {\wv} heavily outperforms {\mj} on non-native speech settings.

%%%%%%%%%%%%%%%%%%%%%%%%%%%%%%%%%%%%%%%%%%%%%%%%%%%%%%%%%%%%%%%%%%%%%%%%%%%
%%%%%%%%%%%%%%%%%%%%%%%%%%%%%%%%%%%%%%%%%%%%%%%%%%%%%%%%%%%%%%%%%%%%%%%%%%%

\subsection{Pronunciation Features}
\label{subsec:Pronunciation Features}

\begin{table*}[ht]
\begin{tabular}{@{}|l|l|@{}}  \toprule
\textbf{Pronunciation feature} & \textbf{Description} \\ \midrule \midrule
StressedSyllPercent & Relative frequency of stressed syllables in percent\\ \midrule
StressDistanceSyllMean & Mean distance between stressed syllables in syllables\\ \midrule
StressDistanceMean & Mean distance between stressed syllables in seconds\\ \midrule
vowelPercentage & Percentage of speech that consists of vowels\\ \midrule
consonantPercentage & Percentage of speech that consists of consonants\\ \midrule
vowelDurationSD & Standard Deviation of vocalic segments\\ \midrule
consonantDurationSD & Standard Deviation of consonantal segments\\ \midrule
syllableDurationSD & Standard Deviation of syllable segments\\ \midrule
vowelSDNorm & Standard Deviation of vowel segments divided by mean\\ & length of vowel segments\\ \midrule
consonantSDNorm & Standard Deviation of consonantal segments divided by\\ & mean length of consonant segments\\ \midrule
syllableSDNorm & Standard Deviation of syllable segments divided by mean\\ & length of syllable segments\\ \midrule
vowelPVINorm & Raw Pairwise Variability Index for vocalic segments\\ \midrule
consonantPVINorm & Raw Pairwise Variability Index for consonantic segments\\ \midrule
syllablePVINorm & Raw Pairwise Variability Index for syllable segments\\ \midrule
\bottomrule
\end{tabular}
\caption{ \label{table: pronunciation feature extraction}Pronunciation feature extraction algorithms and these can extracted easily using the libraries- numpy, textgrids, operator, re, itertools and counter. 
}
\end{table*}

Similar to fluency features, we are the first to probe pronunciation features in speech. The intelligibility, perceived comprehensibility, and accentedness of speech are impacted by phonemic errors \cite{derwing1997accent}. Segmental pronunciation is judged based on the amount of listener effort with lower being the better. Hence, we probe the models for the following pronunciation characteristic features - the percentage, standard deviation, duration and Normalized Pairwise Variability Index (PVI) for vowels (\emph{vowelPercentage, vowelDurationSD, vowelSDNorm, vowelIPVINorm}), consonants (\emph{consontantPercentage, consontantDurationSD, consonantSDNorm, consonantIPVINorm}), and syllables (\emph{syllableDurationSD, syllableSDNorm, syllablePVINorm}). We also study the presence of stress with the characteristic features of stress syllables distance mean (\emph{stressDistanceMean}), and stress distance mean (\emph{stressDistanceSyllMean}).
 
\begin{figure*}
  %\noindent
  \includegraphics[scale=0.5]{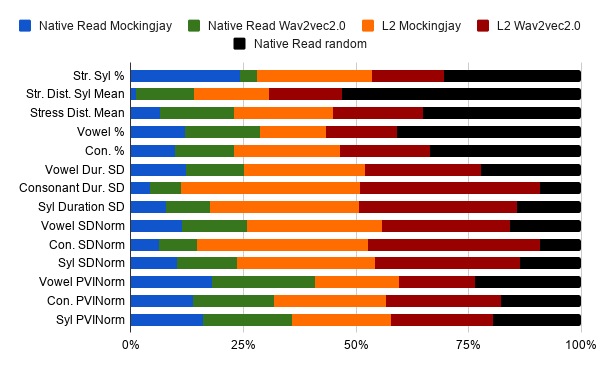}
  \caption{\label{Pronunciation}Performance of each pronunciation feature (on the y-axis) relative to the the performance of random embeddings on L2 Arctic data features \emph{(loss*100/l2\_random\_loss)} on the x-axis where loss values are that of MSE.}
\end{figure*}

\textbf{Native Read Speech:} Figures~\ref{figure:Comparison of w2v and MJ on native read speech}(a2) and (b2)\footnote{Refer Tables \ref{Pron_W} and \ref{Pron_M} of Appendix for the loss values} show the results for probing pronunciation features on wav2vec2.0 and Mockingjay with the Librispeech data. These features are learnt best by the last layers in Mockingjay. Wav2vec2.0 learns these features the most in the 6\textsuperscript{th} to 8\textsuperscript{th} layers amongst its $12$ layers. Mockingjay performs better for pronunciation-based features than wav2vec2.0 by $30.4\%$ in the final layer embeddings. Comparing the minimum loss layers for both models, the difference is $16.19\%$ in favor of Mockingjay. %We also find that features like \emph{syllableSDNorm} and \emph{vowelIPVINorm} perform approximately equally on all the Mockingjay layers but this is not so for wav2vec2.0. 

\textbf{Non-native Speech:}
Mockingjay follows the same pattern for L2 Arctic dataset as for 
the Librispeech dataset. It learns these features better in the last layers. However, for wav2vec2.0, the layers learning each of these pronunciation features are more spread out across the initial layers of the second half of the model. Wav2vec2.0 outperforms Mockingjay but the differences here are reduced to $8.9\%$ in the end layer and $2.20\%$ in the best performing layer. This pattern follows the non-native speech performance of {\wv} and {\mj} seen with audio and fluency features. Here too, the performance difference between {\wv} and {\mj} widens when compared to the native speech scenario.

\subsection{Feature Extractor Module of {\wv}}
\begin{figure*}
  %\noindent
  \includegraphics[scale=0.5]{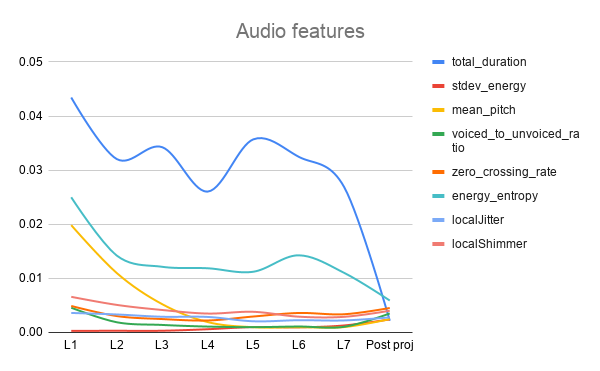}
  \caption{\label{feature_extractor}Performance of audio features on the various layers of Feature Extractor}
\end{figure*}

As shown in Figure~\ref{fig:wv arch}, {\wv} has 7 convolutional layers before the transformer encoder block. The authors call it the ``feature extractor'' of \wv. While in the computer vision community, it has been shown that subsequent layers of a CNN architecture look for higher level features, in the speech community this question has largely been left unaddressed \cite{erhan2010understanding,gong2018deep}. We find that there is a uniform increase in performance of the subsequent CNN layers for all feature types (audio, fluency, and pronunciation) and there is no difference between any features with respect to ``high-level'' or ``low-level''. Figure~\ref{feature_extractor} shows this behavior for audio features(which are supposed to be best learnt by feature extractor of audio transformer). The CNN layers faithfully extract all the features and show minimum loss at the seventh layer or the post-projection layers.

%%%%%%%%%%%%%%%%%%%%%%%%%%%%%%%%%%%%%%%%%%%%%%%%%%%%%%%%%%%%%%%%%%%%%%%%%%%
%%%%%%%%%%%%%%%%%%%%%%%%%%%%%%%%%%%%%%%%%%%%%%%%%%%%%%%%%%%%%%%%%%%%%%%%%%%
\section{Can Audio Models Read Too?}
\label{sec:Can Audio Models Read Too?}
Speech combines the text and the audio parts of the language. Conventionally, the audio community (which also deals with speech) has been more involved with signal sciences while the NLP community has dealt with the text part of speech while ignoring audio. This approach is suboptimal. However, due to the impressive performance of self-supervised transformers in every domain, there is a newfound interest in learning task-independent representations. Concurrently, there is also an interest in learning how these representations are working. Therefore, we probe to check whether the self-supervised audio transformers on account of their self-supervision tasks have accumulated some knowledge present in the text as well. With this motivation, we probe the audio transformer representations for surface (\S\ref{sec:surface level features}), syntax (\S\ref{sec:syntax level features}) and semantic (\S\ref{sec:semantic level features}) knowledge. For reference, we compare them with BERT based text-only embeddings. We use random embeddings as baseline. We do the experiments for four speech types (native read, native spontaneous, non-native, and artificial speech).

While the surface features measure the non-linguistic surface knowledge of the encoders, syntax features measure the syntax based linguistic properties. Conneau \textit{et al.} \cite{conneau2018you} include features such as sentence length and word content in surface features and syntax tree depth in syntax feature. The other category of features we measure are semantics features in which we include number of objects and subjects \cite{conneau2018you}.

\begin{table*}[ht]
\begin{tabular}{@{}|l|l|@{}}  \toprule
\textbf{Text feature} & \textbf{Description}\\ \midrule \midrule
\textbf{Surface Features} \\ \midrule
Unique word count & Total count of unique words(Ignore words of length 3 or smaller)\\ \midrule
Word Complexity & Sum of word complexities for all words in text given by annotators\\ \midrule

\textbf{Semantic Features} \\ \midrule
Total adjectives & Total count of adjectives \\ \midrule
Total adverbs & Total count of adverbs\\ \midrule
Total nouns & Total count of nouns\\ \midrule
Total verbs & Total count of verbs\\ \midrule
Total pronouns & Total count of pronouns\\ \midrule
Total conjunction & Total count of conjunction\\ \midrule
Total conjunction & Total count of conjunction\\ \midrule
Number of subject & Total count of subject\\ \midrule
Number of Object & Total count of direct objects\\ \midrule
Tense & Classification of main clause verb into present or past tense\\ \midrule

\textbf{Syntax Feature} \\ \midrule
Depth of syntax tree & Depth of syntax tree of the text\\ 
\bottomrule
\end{tabular}
\caption{ \label{table: Text feature extraction}Text feature extraction algorithms extracted using nltk and numpy libraries
}
\end{table*}

%%%%%%%%%%%%%%%%%%%%%%%%%%%%%%%%%%%%%%%%%%%%%%%%%%%%%%%%%%%%%%%%%%%%%%%%%%%
%%%%%%%%%%%%%%%%%%%%%%%%%%%%%%%%%%%%%%%%%%%%%%%%%%%%%%%%%%%%%%%%%%%%%%%%%%%
\subsection{Surface Level Features}
\label{sec:surface level features}
Surface level features measure the surface properties of sentences. No linguistic knowledge is required for these features. They can be measured by just looking at the tokens \cite{conneau2018you}. We include the following features - \emph{unique word count} and the average word complexity (\emph{Word Complexity}) since the lexical diversity of spoken speech is an important metric to evaluate its quality \cite{read2006investigation}.

\textbf{Native Read Speech}:
When compared on LibriSpeech, surface-based features are learnt better by Mockingjay than wav2vec2.0 by $9.99\%$ and (b3)\footnote{Refer Tables~\ref{Vocab_W} and \ref{Vocab_M} of Appendix for the loss values}. These features are learnt best in the intermediate layers in wav2vec2.0 and initial layers in Mockingjay. From the results, we observe that the text understanding of both models becomes increasingly diffused as we go towards the later layers. However, wav2vec2.0 outperforms Mockingjay by $3.01\%$ in the final layer. A contributing factor to these observations is the learning of surface features by Mockingjay in the initial layers while, {\wv} learns it best in the middle layers.

\textbf{Non-native Speech}:
For L2 arctic data, again {\wv} best learns the surface features in the middle layers but for mockingjay, no particular pattern is observed. The difference widens to $38.41\%$ on the end layers and $18.96\%$  on the minimum loss layer in favour of {\wv}.

\textbf{Native Spontaneous Speech}:{\mj} learns best in the initial layers like in the case with native read speech meanwhile, {\wv} performs best in the lower middle(7-11) layers. The difference increases to $141.42\%$ for native spontaneous speech on the final layer and $132.44\%$ on the best performing layer.

%%%%%%%%%%%%%%%%%%%%%%%%%%%%%%%%%%%%%%%%%%%%%%%%%%%%%%%%%%%%%%%%%%%%%%%%%%%

\subsection{Semantic Level Features}
\label{sec:semantic level features}
The relationship between the words spoken and our comprehension of that spoken content falls into the domain of semantics. To produce meaning in a sentence, it is almost necessary for it to have a subject and a direct object that the subject addresses. The \emph{number of subjects, number of direct objects and total nouns, pronouns, adverbs, adjectives, verbs, conjunction, and determiners} are hence in our set of features to evaluate the spoken content. \cite{conneau2018you,jawahar-etal-2019-bert}. We also probe for the tense(past or present) and it is framed as a classification task unlike the rest which are regression tasks so the result for tense are separately mentioned.

\begin{figure*}
  %\noindent
  \includegraphics[scale=0.5]{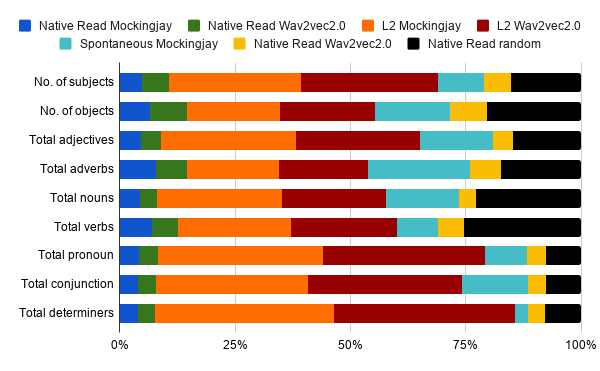}
  \caption{\label{Vocabulary}Performance of each semantic level feature (on the y-axis) relative to the the performance of random embeddings on L2 Arctic data features \emph{(loss*100/l2\_random\_loss)} on the x-axis where loss values are that of MSE.}
\end{figure*}

\textbf{Native Read Speech}:
wav2vec2.0 performs better in this setting by $4.173\%$ and $5.29\%$ on the minimum loss layer. Like the surface features, the pattern followed by the layers in learning is same for semantic features. Mockingjay learns them best in initial layers while wav2vec2.0 in the intermediate layers. For tense too, {\wv} best performs with $75.04\%$ accuracy in the seventh layer where {\mj} performs with $56.99\%$ in the last layer.

\textbf{Non-native Speech}:The same pattern as surface features in the non-native setting is followed by both the transformers. {\mj} does not follow a clear pattern but {\wv} performs best in the middle layers. While wav2vec2.0 outperforms Mockingjay be $7.36\%$ on minimum layer loss for L2 speech, the margin decreases to $3.26\%$ on the end layer. Accuracy for tense is $57.95\%$ for {\wv} and $52.27\%$ for {\mj} on 5th and 9th layer respectively.

\textbf{Native Spontaneous Speech}:{\mj} does not concentrate its learning in any particular layer but {\wv} performs best in the second half of the transformer layers. {\wv} performs better by $9.83\%$ for native spontaneous speech on the best performing layer and $8.06\%$ on the final layer. Again for tense, the accuracy is $65.79\%$ on {\wv} and $57.89\%$ on {\mj}.
%%%%%%%%%%%%%%%%%%%%%%%%%%%%%%%%%%%%%%%%%%%%%%%%%%%%%%%%%%%%%%%%%%%%%%%%%%%
%%%%%%%%%%%%%%%%%%%%%%%%%%%%%%%%%%%%%%%%%%%%%%%%%%%%%%%%%%%%%%%%%%%%%%%%%%%
\subsection{Syntax Level Features}
\label{sec:syntax level features}

Syntax is the key component of the grammatical structure of a sentence, which in turn is a key component of the communicative competence \cite{canale1980theoretical}. We use the \emph{depth of the syntax tree} constructed from the sentences spoken in each sound clip as a feature to evaluate the syntax content \cite{conneau2018you,jawahar-etal-2019-bert,kumar2019get}.

\textbf{Native Read Speech}:In this setting as well, Mockingjay performs better than wav2vec2.0 by $38.64\%$ on the best performing layer and by $21.5\%$ on the final layer. The final layer captures this feature best for {\wv} and the initial for {\mj}, which explains the decrease in percentage difference for the final layer. 

\textbf{Non-native Speech}:
{\wv} performs better on minimum layer loss by $15.89\%$ and $30.92\%$ on the final layer. {\wv} learns best on eight layer and {\mj} learns best on fourth layer.

\textbf{Native Spontaneous Speech}:
%%%%%%%%%%%%%%%%%%%%%%%%%%%%%%%%%%%%%%%%%%%%%%%%%%%%%%%%%%%%%%%%%%%%%%%%%%%
%%%%%%%%%%%%%%%%%%%%%%%%%%%%%%%%%%%%%%%%%%%%%%%%%%%%%%%%%%%%%%%%%%%%%%%%%%%
\subsection{Feature Extractor Module of {\wv}} 
\label{Feature Extractor Module of wav2vec2.0 text features}
The pattern observed in the feature extractor module for these surface level features is the same as that of audio features with minimum losses seen in the post projection layer. However, the value of the minimum loss in this layer is less than that of the transformer module in wav2vec2.0. This gives some intuition for the better performance of Mockingjay since the Transformer is unable to capture the features or unlearns the presented vocabulary features.

%%%%%%%%%%%%%%%%%%%%%%%%%%%%%%%%%%%%%%%%%%%%%%%%%%%%%%%%%%%%%%%%%%%%%%%%%%%
%%%%%%%%%%%%%%%%%%%%%%%%%%%%%%%%%%%%%%%%%%%%%%%%%%%%%%%%%%%%%%%%%%%%%%%%%%%
\subsection{Comparison with BERT} 
\begin{table*}[!htbp]
\small
\centering
\begin{tabular}{|l|l|l|l|l|}
\hline
\textbf{Dataset} & \textbf{Model} & \textbf{Semantic} & \textbf{Syntax} & \textbf{Surface}   \\\hline
Native Read Speech & wav2vec2.0 &  -43.62\%, -40.23\% & -56.90\%, -67.15\% & -59.53\%, -51.31\%\\
& Mockingjay & -41.78\%, -39.55\% & -73.57\%, -74.21\%  & -52.20\%, -47.99\%\\ \hline
Non-native Read Speech & wav2vec2.0 & 15.88\%, 7.35\% & 59.05\%, 30.33\% & 78.27\%, -3.94\%\\
& Mockingjay & 24.30\%, 11.14\% & 79.72\%, 33.97\% & 121.71\%, 21.30\% \\ \hline
Wikipedia TTS & wav2vec2.0 & 10.22\%, -2.29\%& -34.55\%, 3.83\% & 17.70\%, -7.58\%\\ 
& Mockingjay  & 13.87\%, -0.49\% & -47.70\%, 7.68\% & 4.04\%, 21.90\% \\ \hline
\end{tabular}
\vspace{1 mm}
\caption{\label{table:text-feature-compared-with-bert} \small Table for comparison of the performance of BERT with wav2vec2.0 and Mockingjay on text features. The two values mentioned per cell indicate the relative minimum loss across all the model layers and the relative end layer losses when compared with the corresponding values for BERT. The values shown are an average across all features of a particular category with the relative performance calculated as $(model\_loss - bert\_loss)*100\%/bert\_loss$. %\cy{Why the losses are represented with percentage?}
}
\vspace{-4 mm}
\end{table*}

When we compare the performance of audio-transformer models with BERT (Table~\ref{table:text-feature-compared-with-bert}) on the native read speech, we observe that on an average, both wav2vec2.0 and Mockingjay perform better than BERT by 43.62\% and 41.78\% on semantic features, 56.90\% and 73.57\% on syntactic features and 59.53\% and 52.20\% on semantic features respectively. These results are surprising since none of the speech transformer models was trained with text objective functions. We hypothesize that this could be due to differences in the train set of the three models. LibriSpeech is the train-set for both the speech-transformer models where as Wikipedia is the train-set for BERT. To confirm this, we test the performance of the three models on text features extracted from Wikipedia and native spontaneous speech datasets. These datasets provide us with a comprehensive comparison. While on one hand, Wikipedia is the train-set for BERT, and the text features from Wikipedia articles are very different from LibriSpeech, on the other, non-native read speech dataset can be considered out-of-domain for both the speech transformer models and BERT.

For the first part, we convert 2000 random sentences from Wikipedia articles to speech by using Google's text-to-speech API \cite{gtts}. We made sure that the audios constructed had similar lengths as those of LibriSpeech. The audios obtained were then passed through both the speech Transformer models and the layers were then probed. On this synthetic dataset, for the semantic features, BERT outperforms both the models by more than 10\% when compared on minimum loss across all the layers. However, by the end layers, both the models learn the features well and the performance difference between BERT and audio-transformer models reduces greatly ($2.29\%$ and $0.49\%$ difference for semantic features, $3.83\%$ and $7.68\%$ for syntax and $7.58\%$ and $21.90\%$ for surface features). These results are motivating since this means that embeddings of audio Transformer captures not only audio, fluency and pronunciation features, but also textual features to a large extent.

Next, we use the CMU L2 Arctic dataset. Table~\ref{table:text-feature-compared-with-bert} presents the results for all the experiments. Here the results are the most different from the previous ones. For the semantic, syntax and surface features, BERT outperforms both the models by more than $15\%$. This result when compared with Wikipedia TTS and native read speech implies that the audio models capture text features for native speakers in `cleaner settings' but they are not able to work in not-so controlled environments. Therefore, in a general setting, BERT text embeddings combined with audio embeddings can capture all the speech features adequately.

%%%%%%%%%%%%%%%%%%%%%%%%%%%%%%%%%%%%%%%%%%%%%%%%%%%%%%%%%%%%%%%%%%%%%%%%%%%
%%%%%%%%%%%%%%%%%%%%%%%%%%%%%%%%%%%%%%%%%%%%%%%%%%%%%%%%%%%%%%%%%%%%%%%%%%%
\section{Effect on Downstream Tasks}
\label{sec:Effect on Downstream Tasks}
We wanted to evaluate our findings which show that different layers of the models capture different features and see its impact on downstream tasks. To this end, we perform two representative tasks: speaker recognition on Voxceleb \cite{nagrani2017voxceleb} (which uses audio features primarily), and phone classification on LibriSpeech (which uses pronunciation features). 

For speaker recognition, we randomly pick $10$ speakers with $50$ audios each in the train-set and $10$ in the test-set. For phone classification, we use the libri-clean-100 and libri-cleanTest splits. We build a 4-layer linear classifier with dimensions $756, 512, 256, 10$ with Adam optimizer and a learning rate of $0.01$. Hidden layers have ReLU activation function and the third layer also has dropout. We perform the tasks using the best performing, final, and weighted average of all layer embeddings of the transformer models as input. 

\begin{table}[!htbp]
\small
\centering

\begin{tabular}{|l|l|l|l|}
\hline
 & \textbf{Best} & \textbf{Last} & \textbf{Wtd Avg}  \\\hline
{\wv} & 91\%/81\% & 31\%/70\% & 87\%/77\% \\\hline
{\mj} & 10\%/83\% & 32\%/83\% & 26\%/79\% \\\hline
\end{tabular}

\caption{\label{table:perf-downstream-tasks-wv-mj} \small Comparison of the performance of {\wv} and {\mj} on (speaker recognition/phone classification) tasks. Here \textit{best} denotes best performing layer, \textit{i.e.}, first for audio features and sixth for pronunciation for {\wv}.
}

\end{table}
Results for both the tasks are given in Table~\ref{table:perf-downstream-tasks-wv-mj}. The results are consistent with those found for audio (\S\ref{subsec:Audio Features}) and pronunciation features (\S\ref{subsec:Pronunciation Features}).
%%%%%%%%%%%%%%%%%%%%%%%%%%%%%%%%%%%%%%%%%%%%%%%%%%%%%%%%%%%%%%%%%%%%%%%%%%%
%%%%%%%%%%%%%%%%%%%%%%%%%%%%%%%%%%%%%%%%%%%%%%%%%%%%%%%%%%%%%%%%%%%%%%%%%%%

\section{Other Related Work}
\label{sec:related work}
We already covered closely related work on attribution in Sections~1 and \ref{sec:Brief Overview Of The Probed Models}. We mention other related work.

\textbf{Audio Probing:} In the domain of speech processing, probes have been carried out on feature vectors, neural networks like RNN or DNN, end-to-end ASR systems or Audio-visual models. In \cite{raj2019probing}, probing on x-vectors which are trained solely to predict the speaker label revealed they also contain incidental information about the transcription, channel, or meta-information about the utterance. Probing the Music Information Retrieval(MIR) prediction through Local Interpretable Model-Agnostic Explanations (LIME) by using AudioLIME \cite{haunschmid2020audiolime} helped interpret MIR for the first time. \cite{nagamine2015exploring} analyses a DNN for phoneme recognition, both at single node and poplation level. Further research on interpretation of the role of non-linear activation of the nodes of a sigmoid DNN built for phoneme recognition task is done in \cite{nagamine2016role}. Research has also been done to address why LSTMs work well as a sequence model for  statistical parametric speech synthesis \cite{wu2016investigating}. Several other studies have been conducted to interpret the correlation between audio and image structures for audio-visual tasks  \cite{alishahi2017encoding,drexler2017analysis,harwath2017learning}. Even for Deep ASR models, efforts have been made to comprehend the hidden and learned representations \cite{belinkov2017analyzing,elloumi2018analyzing}. However, probing of representation learning audio transformers is yet unexplored.

\textbf{Text Probing:}
The field of natural language processing has seen numerous efforts in understanding the inner working of large-scale transformers, especially BERT \cite{jawahar-etal-2019-bert,cui2020does,ramnath2020towards}. \citet{jawahar-etal-2019-bert} probe each of the different layers of BERT to find which layers best learn the phrase-level information, linguistic information and the long-distance dependencies. The results showed what role each layer played and the study concluded that the middle layers learnt the syntactic features and the higher levels learnt the semantic features and that the deeper layers are needed for long-distance dependencies while the initial layers capture the phrase-level information. 
%You can leave this. I will complete it.Okay XXX
%%%%%%%%%%%%%%%%%%%%%%%%%%%%%%%%%%%%%%%%%%%%%%%%%%%%%%%%%%%%%%%%%%%%%%%%%%%
%%%%%%%%%%%%%%%%%%%%%%%%%%%%%%%%%%%%%%%%%%%%%%%%%%%%%%%%%%%%%%%%%%%%%%%%%%%

\section{Conclusion}
Speech transformer models, while still being new, have shown state-of-the-art performance on various downstream tasks. We probe two such models, wav2vec2.0 and Mockingjay, to understand what they learn. We probe the models on a wide range of features including audio, fluency, suprasegmental pronunciation, and text-based characteristics. For each category of features, we identify a learning pattern over each model and its layers. We find that {\wv} outperforms {\mj} on audio and fluency features but underperforms on pronunciation features. Furthermore, we compare BERT with the audio models with text features and find that the audio models surprisingly outperform BERT in cleaner, controlled settings of native speech, 
%(native read speech and synthetic speech in native voice) 
but are not able to perform in an uncontrolled environment such as of spontaneous speech and non-native speech. %We show our results on a variety of settings including native, non-native, read, spontaneous and synthetic speech datasets.
\begin{acks}
To Robert, for the bagels and explaining CMYK and color spaces.
\end{acks}

\bibliographystyle{ACM-Reference-Format}
\bibliography{sample-base}


\begin{thebibliography}{67}

%%% the Web address.

\ifx \showCODEN    \undefined \def \showCODEN     #1{\unskip}     \fi
\ifx \showDOI      \undefined \def \showDOI       #1{#1}\fi
\ifx \showISBNx    \undefined \def \showISBNx     #1{\unskip}     \fi
\ifx \showISBNxiii \undefined \def \showISBNxiii  #1{\unskip}     \fi
\ifx \showISSN     \undefined \def \showISSN      #1{\unskip}     \fi
\ifx \showLCCN     \undefined \def \showLCCN      #1{\unskip}     \fi
\ifx \shownote     \undefined \def \shownote      #1{#1}          \fi
\ifx \showarticletitle \undefined \def \showarticletitle #1{#1}   \fi
\ifx \showURL      \undefined \def \showURL       {\relax}        \fi
% invisible to TeX
\providecommand\bibfield[2]{#2}
\providecommand\bibinfo[2]{#2}
\providecommand\natexlab[1]{#1}
\providecommand\showeprint[2][]{arXiv:#2}

\bibitem[\protect\citeauthoryear{Alishahi, Barking, and Chrupa{\l}a}{Alishahi
  et~al\mbox{.}}{2017}]%
        {alishahi2017encoding}
\bibfield{author}{\bibinfo{person}{Afra Alishahi}, \bibinfo{person}{Marie
  Barking}, {and} \bibinfo{person}{Grzegorz Chrupa{\l}a}.}
  \bibinfo{year}{2017}\natexlab{}.
\newblock \showarticletitle{Encoding of phonology in a recurrent neural model
  of grounded speech}. In \bibinfo{booktitle}{\emph{Proceedings of the 21st
  Conference on Computational Natural Language Learning (CoNLL 2017)}}.
  \bibinfo{pages}{368--378}.
\newblock


\bibitem[\protect\citeauthoryear{Ardila, Branson, Davis, Henretty, Kohler,
  Meyer, Morais, Saunders, Tyers, and Weber}{Ardila et~al\mbox{.}}{2019}]%
        {ardila2019common}
\bibfield{author}{\bibinfo{person}{Rosana Ardila}, \bibinfo{person}{Megan
  Branson}, \bibinfo{person}{Kelly Davis}, \bibinfo{person}{Michael Henretty},
  \bibinfo{person}{Michael Kohler}, \bibinfo{person}{Josh Meyer},
  \bibinfo{person}{Reuben Morais}, \bibinfo{person}{Lindsay Saunders},
  \bibinfo{person}{Francis~M Tyers}, {and} \bibinfo{person}{Gregor Weber}.}
  \bibinfo{year}{2019}\natexlab{}.
\newblock \showarticletitle{Common voice: A massively-multilingual speech
  corpus}.
\newblock \bibinfo{journal}{\emph{arXiv preprint arXiv:1912.06670}}
  (\bibinfo{year}{2019}).
\newblock


\bibitem[\protect\citeauthoryear{Baevski, Schneider, and Auli}{Baevski
  et~al\mbox{.}}{2019}]%
        {baevski2019vq}
\bibfield{author}{\bibinfo{person}{Alexei Baevski}, \bibinfo{person}{Steffen
  Schneider}, {and} \bibinfo{person}{Michael Auli}.}
  \bibinfo{year}{2019}\natexlab{}.
\newblock \showarticletitle{vq-wav2vec: Self-Supervised Learning of Discrete
  Speech Representations}. In \bibinfo{booktitle}{\emph{International
  Conference on Learning Representations}}.
\newblock


\bibitem[\protect\citeauthoryear{Baevski, Zhou, Mohamed, and Auli}{Baevski
  et~al\mbox{.}}{2020}]%
        {baevski2020wav2vec}
\bibfield{author}{\bibinfo{person}{Alexei Baevski}, \bibinfo{person}{Henry
  Zhou}, \bibinfo{person}{Abdelrahman Mohamed}, {and} \bibinfo{person}{Michael
  Auli}.} \bibinfo{year}{2020}\natexlab{}.
\newblock \showarticletitle{wav2vec 2.0: A framework for self-supervised
  learning of speech representations}.
\newblock \bibinfo{journal}{\emph{arXiv preprint arXiv:2006.11477}}
  (\bibinfo{year}{2020}).
\newblock


\bibitem[\protect\citeauthoryear{Barth, Tolar, Fletcher, and Francis}{Barth
  et~al\mbox{.}}{2014}]%
        {barth2014effects}
\bibfield{author}{\bibinfo{person}{Amy~E Barth}, \bibinfo{person}{Tammy~D
  Tolar}, \bibinfo{person}{Jack~M Fletcher}, {and} \bibinfo{person}{David
  Francis}.} \bibinfo{year}{2014}\natexlab{}.
\newblock \showarticletitle{The effects of student and text characteristics on
  the oral reading fluency of middle-grade students.}
\newblock \bibinfo{journal}{\emph{Journal of Educational Psychology}}
  \bibinfo{volume}{106}, \bibinfo{number}{1} (\bibinfo{year}{2014}),
  \bibinfo{pages}{162}.
\newblock


\bibitem[\protect\citeauthoryear{Belinkov, Durrani, Dalvi, Sajjad, and
  Glass}{Belinkov et~al\mbox{.}}{2017}]%
        {belinkov2017neural}
\bibfield{author}{\bibinfo{person}{Yonatan Belinkov}, \bibinfo{person}{Nadir
  Durrani}, \bibinfo{person}{Fahim Dalvi}, \bibinfo{person}{Hassan Sajjad},
  {and} \bibinfo{person}{James Glass}.} \bibinfo{year}{2017}\natexlab{}.
\newblock \showarticletitle{What do neural machine translation models learn
  about morphology?}
\newblock \bibinfo{journal}{\emph{arXiv preprint arXiv:1704.03471}}
  (\bibinfo{year}{2017}).
\newblock


\bibitem[\protect\citeauthoryear{Belinkov and Glass}{Belinkov and
  Glass}{2017}]%
        {belinkov2017analyzing}
\bibfield{author}{\bibinfo{person}{Yonatan Belinkov} {and}
  \bibinfo{person}{James Glass}.} \bibinfo{year}{2017}\natexlab{}.
\newblock \showarticletitle{Analyzing hidden representations in end-to-end
  automatic speech recognition systems}. In \bibinfo{booktitle}{\emph{Advances
  in Neural Information Processing Systems}}. \bibinfo{pages}{2441--2451}.
\newblock


\bibitem[\protect\citeauthoryear{Boersma and Weenink}{Boersma and
  Weenink}{2021}]%
        {praat}
\bibfield{author}{\bibinfo{person}{Paul Boersma} {and} \bibinfo{person}{David
  Weenink}.} \bibinfo{year}{2021}\natexlab{}.
\newblock \bibinfo{title}{{P}raat: doing phonetics by computer [{C}omputer
  program]}.
\newblock \bibinfo{howpublished}{Version 6.1.38, retrieved 2 January 2021
  \url{http://www.praat.org/}}.
\newblock


\bibitem[\protect\citeauthoryear{Canale and Swain}{Canale and Swain}{1980}]%
        {canale1980theoretical}
\bibfield{author}{\bibinfo{person}{Michael Canale} {and}
  \bibinfo{person}{Merrill Swain}.} \bibinfo{year}{1980}\natexlab{}.
\newblock \showarticletitle{Theoretical bases of communicative approaches to
  second language teaching and testing}.
\newblock \bibinfo{journal}{\emph{Applied linguistics}} \bibinfo{volume}{1},
  \bibinfo{number}{1} (\bibinfo{year}{1980}), \bibinfo{pages}{1--47}.
\newblock


\bibitem[\protect\citeauthoryear{Chi, Chung, Wu, Hsieh, Li, and Lee}{Chi
  et~al\mbox{.}}{2020}]%
        {chi2020audio}
\bibfield{author}{\bibinfo{person}{Po-Han Chi}, \bibinfo{person}{Pei-Hung
  Chung}, \bibinfo{person}{Tsung-Han Wu}, \bibinfo{person}{Chun-Cheng Hsieh},
  \bibinfo{person}{Shang-Wen Li}, {and} \bibinfo{person}{Hung-yi Lee}.}
  \bibinfo{year}{2020}\natexlab{}.
\newblock \showarticletitle{Audio ALBERT: A Lite BERT for Self-supervised
  Learning of Audio Representation}.
\newblock \bibinfo{journal}{\emph{arXiv preprint arXiv:2005.08575}}
  (\bibinfo{year}{2020}).
\newblock


\bibitem[\protect\citeauthoryear{Choi, Fazekas, Cho, and Sandler}{Choi
  et~al\mbox{.}}{2017}]%
        {choi2017tutorial}
\bibfield{author}{\bibinfo{person}{Keunwoo Choi}, \bibinfo{person}{Gy{\"o}rgy
  Fazekas}, \bibinfo{person}{Kyunghyun Cho}, {and} \bibinfo{person}{Mark
  Sandler}.} \bibinfo{year}{2017}\natexlab{}.
\newblock \showarticletitle{A tutorial on deep learning for music information
  retrieval}.
\newblock \bibinfo{journal}{\emph{arXiv preprint arXiv:1709.04396}}
  (\bibinfo{year}{2017}).
\newblock


\bibitem[\protect\citeauthoryear{Conneau, Kruszewski, Lample, Barrault, and
  Baroni}{Conneau et~al\mbox{.}}{2018}]%
        {conneau2018you}
\bibfield{author}{\bibinfo{person}{Alexis Conneau}, \bibinfo{person}{Germ{\'a}n
  Kruszewski}, \bibinfo{person}{Guillaume Lample}, \bibinfo{person}{Lo{\"\i}c
  Barrault}, {and} \bibinfo{person}{Marco Baroni}.}
  \bibinfo{year}{2018}\natexlab{}.
\newblock \showarticletitle{What you can cram into a single vector: Probing
  sentence embeddings for linguistic properties}.
\newblock \bibinfo{journal}{\emph{arXiv preprint arXiv:1805.01070}}
  (\bibinfo{year}{2018}).
\newblock


\bibitem[\protect\citeauthoryear{Cui, Cheng, Wu, and Zhang}{Cui
  et~al\mbox{.}}{2020}]%
        {cui2020does}
\bibfield{author}{\bibinfo{person}{Leyang Cui}, \bibinfo{person}{Sijie Cheng},
  \bibinfo{person}{Yu Wu}, {and} \bibinfo{person}{Yue Zhang}.}
  \bibinfo{year}{2020}\natexlab{}.
\newblock \showarticletitle{Does BERT Solve Commonsense Task via Commonsense
  Knowledge?}
\newblock \bibinfo{journal}{\emph{arXiv preprint arXiv:2008.03945}}
  (\bibinfo{year}{2020}).
\newblock


\bibitem[\protect\citeauthoryear{De~Jong and Wempe}{De~Jong and Wempe}{2009}]%
        {de2009praat}
\bibfield{author}{\bibinfo{person}{Nivja~H De~Jong} {and} \bibinfo{person}{Ton
  Wempe}.} \bibinfo{year}{2009}\natexlab{}.
\newblock \showarticletitle{Praat script to detect syllable nuclei and measure
  speech rate automatically}.
\newblock \bibinfo{journal}{\emph{Behavior research methods}}
  \bibinfo{volume}{41}, \bibinfo{number}{2} (\bibinfo{year}{2009}),
  \bibinfo{pages}{385--390}.
\newblock


\bibitem[\protect\citeauthoryear{Derwing and Munro}{Derwing and Munro}{1997}]%
        {derwing1997accent}
\bibfield{author}{\bibinfo{person}{Tracey~M Derwing} {and}
  \bibinfo{person}{Murray~J Munro}.} \bibinfo{year}{1997}\natexlab{}.
\newblock \showarticletitle{Accent, comprehensibility and intelligibility:
  Evidence from four L1s}.
\newblock \bibinfo{journal}{\emph{Studies in Second Language Acquisition}}
  \bibinfo{volume}{19}, \bibinfo{number}{1} (\bibinfo{year}{1997}),
  \bibinfo{pages}{1--16}.
\newblock


\bibitem[\protect\citeauthoryear{Devlin, Chang, Lee, and Toutanova}{Devlin
  et~al\mbox{.}}{2019}]%
        {devlinetal2019bert}
\bibfield{author}{\bibinfo{person}{Jacob Devlin}, \bibinfo{person}{Ming-Wei
  Chang}, \bibinfo{person}{Kenton Lee}, {and} \bibinfo{person}{Kristina
  Toutanova}.} \bibinfo{year}{2019}\natexlab{}.
\newblock \showarticletitle{{BERT}: Pre-training of Deep Bidirectional
  Transformers for Language Understanding}. In
  \bibinfo{booktitle}{\emph{Proceedings of the 2019 Conference of the North
  {A}merican Chapter of the Association for Computational Linguistics: Human
  Language Technologies, Volume 1 (Long and Short Papers)}}.
  \bibinfo{publisher}{Association for Computational Linguistics},
  \bibinfo{address}{Minneapolis, Minnesota}, \bibinfo{pages}{4171--4186}.
\newblock
\urldef\tempurl%
\url{https://doi.org/10.18653/v1/N19-1423}
\showDOI{\tempurl}


\bibitem[\protect\citeauthoryear{Doumbouya, Einstein, and Piech}{Doumbouya
  et~al\mbox{.}}{2021}]%
        {doumbouya2021using}
\bibfield{author}{\bibinfo{person}{Moussa Doumbouya}, \bibinfo{person}{Lisa
  Einstein}, {and} \bibinfo{person}{Chris Piech}.}
  \bibinfo{year}{2021}\natexlab{}.
\newblock \showarticletitle{Using Radio Archives for Low-Resource Speech
  Recognition: Towards an Intelligent Virtual Assistant for Illiterate Users}.
  In \bibinfo{booktitle}{\emph{Proceedings of the AAAI Conference on Artificial
  Intelligence}}, Vol.~\bibinfo{volume}{35}.
\newblock


\bibitem[\protect\citeauthoryear{Drexler and Glass}{Drexler and Glass}{2017}]%
        {drexler2017analysis}
\bibfield{author}{\bibinfo{person}{Jennifer Drexler} {and}
  \bibinfo{person}{James Glass}.} \bibinfo{year}{2017}\natexlab{}.
\newblock \showarticletitle{Analysis of audio-visual features for unsupervised
  speech recognition}. In \bibinfo{booktitle}{\emph{Grounded Language
  Understanding Workshop}}.
\newblock


\bibitem[\protect\citeauthoryear{Durette and Contributors}{Durette and
  Contributors}{2020}]%
        {gtts}
\bibfield{author}{\bibinfo{person}{Pierre~Nicolas Durette} {and}
  \bibinfo{person}{Contributors}.} \bibinfo{year}{2020}\natexlab{}.
\newblock \bibinfo{title}{Google Text to Speech Model}.
\newblock \bibinfo{howpublished}{\url{https://pypi.org/project/gTTS/}}.
\newblock


\bibitem[\protect\citeauthoryear{Elloumi, Besacier, Galibert, and
  Lecouteux}{Elloumi et~al\mbox{.}}{2018}]%
        {elloumi2018analyzing}
\bibfield{author}{\bibinfo{person}{Zied Elloumi}, \bibinfo{person}{Laurent
  Besacier}, \bibinfo{person}{Olivier Galibert}, {and}
  \bibinfo{person}{Benjamin Lecouteux}.} \bibinfo{year}{2018}\natexlab{}.
\newblock \showarticletitle{Analyzing Learned Representations of a Deep ASR
  Performance Prediction Model}. In \bibinfo{booktitle}{\emph{Blackbox NLP
  Workshop and EMLP 2018}}.
\newblock


\bibitem[\protect\citeauthoryear{Erhan, Courville, and Bengio}{Erhan
  et~al\mbox{.}}{2010}]%
        {erhan2010understanding}
\bibfield{author}{\bibinfo{person}{Dumitru Erhan}, \bibinfo{person}{Aaron
  Courville}, {and} \bibinfo{person}{Yoshua Bengio}.}
  \bibinfo{year}{2010}\natexlab{}.
\newblock \showarticletitle{Understanding representations learned in deep
  architectures}.
\newblock \bibinfo{journal}{\emph{Department dInformatique et Recherche
  Operationnelle, University of Montreal, QC, Canada, Tech. Rep}}
  \bibinfo{volume}{1355}, \bibinfo{number}{1} (\bibinfo{year}{2010}).
\newblock


\bibitem[\protect\citeauthoryear{Garofolo, Lamel, Fisher, Fiscus, and
  Pallett}{Garofolo et~al\mbox{.}}{1993}]%
        {garofolo1993darpa}
\bibfield{author}{\bibinfo{person}{John~S Garofolo}, \bibinfo{person}{Lori~F
  Lamel}, \bibinfo{person}{William~M Fisher}, \bibinfo{person}{Jonathan~G
  Fiscus}, {and} \bibinfo{person}{David~S Pallett}.}
  \bibinfo{year}{1993}\natexlab{}.
\newblock \showarticletitle{DARPA TIMIT acoustic-phonetic continous speech
  corpus CD-ROM. NIST speech disc 1-1.1}.
\newblock \bibinfo{journal}{\emph{STIN}}  \bibinfo{volume}{93}
  (\bibinfo{year}{1993}), \bibinfo{pages}{27403}.
\newblock


\bibitem[\protect\citeauthoryear{Giannakopoulos}{Giannakopoulos}{2015}]%
        {giannakopoulos2015pyaudioanalysis}
\bibfield{author}{\bibinfo{person}{Theodoros Giannakopoulos}.}
  \bibinfo{year}{2015}\natexlab{}.
\newblock \showarticletitle{pyaudioanalysis: An open-source python library for
  audio signal analysis}.
\newblock \bibinfo{journal}{\emph{PloS one}} \bibinfo{volume}{10},
  \bibinfo{number}{12} (\bibinfo{year}{2015}), \bibinfo{pages}{e0144610}.
\newblock


\bibitem[\protect\citeauthoryear{Girshick, Donahue, Darrell, and
  Malik}{Girshick et~al\mbox{.}}{2014}]%
        {girshick2014rich}
\bibfield{author}{\bibinfo{person}{Ross Girshick}, \bibinfo{person}{Jeff
  Donahue}, \bibinfo{person}{Trevor Darrell}, {and} \bibinfo{person}{Jitendra
  Malik}.} \bibinfo{year}{2014}\natexlab{}.
\newblock \showarticletitle{Rich feature hierarchies for accurate object
  detection and semantic segmentation}. In
  \bibinfo{booktitle}{\emph{Proceedings of the IEEE conference on computer
  vision and pattern recognition}}. \bibinfo{pages}{580--587}.
\newblock


\bibitem[\protect\citeauthoryear{Gong and Poellabauer}{Gong and
  Poellabauer}{2018}]%
        {gong2018deep}
\bibfield{author}{\bibinfo{person}{Yuan Gong} {and} \bibinfo{person}{Christian
  Poellabauer}.} \bibinfo{year}{2018}\natexlab{}.
\newblock \showarticletitle{How do deep convolutional neural networks learn
  from raw audio waveforms?}
\newblock  (\bibinfo{year}{2018}).
\newblock


\bibitem[\protect\citeauthoryear{Grover, Kumar, Sarin, Vafaee, Hama, and
  Shah}{Grover et~al\mbox{.}}{2020}]%
        {grover2020multi}
\bibfield{author}{\bibinfo{person}{Manraj~Singh Grover}, \bibinfo{person}{Yaman
  Kumar}, \bibinfo{person}{Sumit Sarin}, \bibinfo{person}{Payman Vafaee},
  \bibinfo{person}{Mika Hama}, {and} \bibinfo{person}{Rajiv~Ratn Shah}.}
  \bibinfo{year}{2020}\natexlab{}.
\newblock \showarticletitle{Multi-modal Automated Speech Scoring using
  Attention Fusion}.
\newblock \bibinfo{journal}{\emph{arXiv preprint arXiv:2005.08182}}
  (\bibinfo{year}{2020}).
\newblock


\bibitem[\protect\citeauthoryear{Harwath and Glass}{Harwath and Glass}{2017}]%
        {harwath2017learning}
\bibfield{author}{\bibinfo{person}{David Harwath} {and} \bibinfo{person}{James
  Glass}.} \bibinfo{year}{2017}\natexlab{}.
\newblock \showarticletitle{Learning Word-Like Units from Joint Audio-Visual
  Analysis}. In \bibinfo{booktitle}{\emph{Proceedings of the 55th Annual
  Meeting of the Association for Computational Linguistics (Volume 1: Long
  Papers)}}. \bibinfo{pages}{506--517}.
\newblock


\bibitem[\protect\citeauthoryear{Haunschmid, Manilow, and Widmer}{Haunschmid
  et~al\mbox{.}}{2020}]%
        {haunschmid2020audiolime}
\bibfield{author}{\bibinfo{person}{Verena Haunschmid}, \bibinfo{person}{Ethan
  Manilow}, {and} \bibinfo{person}{Gerhard Widmer}.}
  \bibinfo{year}{2020}\natexlab{}.
\newblock \showarticletitle{audioLIME: Listenable Explanations Using Source
  Separation}.
\newblock \bibinfo{journal}{\emph{arXiv preprint arXiv:2008.00582}}
  (\bibinfo{year}{2020}).
\newblock


\bibitem[\protect\citeauthoryear{Hewitt and Manning}{Hewitt and
  Manning}{2019}]%
        {hewitt2019structural}
\bibfield{author}{\bibinfo{person}{John Hewitt} {and}
  \bibinfo{person}{Christopher~D Manning}.} \bibinfo{year}{2019}\natexlab{}.
\newblock \showarticletitle{A structural probe for finding syntax in word
  representations}. In \bibinfo{booktitle}{\emph{Proceedings of the 2019
  Conference of the North American Chapter of the Association for Computational
  Linguistics: Human Language Technologies, Volume 1 (Long and Short Papers)}}.
  \bibinfo{pages}{4129--4138}.
\newblock


\bibitem[\protect\citeauthoryear{Igras-Cybulska, Zi{\'o}{\l}ko, {\.Z}elasko,
  and Witkowski}{Igras-Cybulska et~al\mbox{.}}{2016}]%
        {igras2016structure}
\bibfield{author}{\bibinfo{person}{Magdalena Igras-Cybulska},
  \bibinfo{person}{Bartosz Zi{\'o}{\l}ko}, \bibinfo{person}{Piotr {\.Z}elasko},
  {and} \bibinfo{person}{Marcin Witkowski}.} \bibinfo{year}{2016}\natexlab{}.
\newblock \showarticletitle{Structure of pauses in speech in the context of
  speaker verification and classification of speech type}.
\newblock \bibinfo{journal}{\emph{EURASIP Journal on Audio, Speech, and Music
  Processing}} \bibinfo{volume}{2016}, \bibinfo{number}{1}
  (\bibinfo{year}{2016}), \bibinfo{pages}{18}.
\newblock


\bibitem[\protect\citeauthoryear{Jadoul, Thompson, and de~Boer}{Jadoul
  et~al\mbox{.}}{2018}]%
        {parselmouth}
\bibfield{author}{\bibinfo{person}{Yannick Jadoul}, \bibinfo{person}{Bill
  Thompson}, {and} \bibinfo{person}{Bart de Boer}.}
  \bibinfo{year}{2018}\natexlab{}.
\newblock \showarticletitle{Introducing {P}arselmouth: A {P}ython interface to
  {P}raat}.
\newblock \bibinfo{journal}{\emph{Journal of Phonetics}}  \bibinfo{volume}{71}
  (\bibinfo{year}{2018}), \bibinfo{pages}{1--15}.
\newblock
\urldef\tempurl%
\url{https://doi.org/10.1016/j.wocn.2018.07.001}
\showDOI{\tempurl}


\bibitem[\protect\citeauthoryear{Jawahar, Sagot, and Seddah}{Jawahar
  et~al\mbox{.}}{2019}]%
        {jawahar-etal-2019-bert}
\bibfield{author}{\bibinfo{person}{Ganesh Jawahar},
  \bibinfo{person}{Beno{\^\i}t Sagot}, {and} \bibinfo{person}{Djam{\'e}
  Seddah}.} \bibinfo{year}{2019}\natexlab{}.
\newblock \showarticletitle{What Does {BERT} Learn about the Structure of
  Language?}. In \bibinfo{booktitle}{\emph{Proceedings of the 57th Annual
  Meeting of the Association for Computational Linguistics}}.
  \bibinfo{publisher}{Association for Computational Linguistics},
  \bibinfo{address}{Florence, Italy}, \bibinfo{pages}{3651--3657}.
\newblock
\urldef\tempurl%
\url{https://doi.org/10.18653/v1/P19-1356}
\showDOI{\tempurl}


\bibitem[\protect\citeauthoryear{Jyothi and Hasegawa-Johnson}{Jyothi and
  Hasegawa-Johnson}{2015}]%
        {jyothi2015improved}
\bibfield{author}{\bibinfo{person}{Preethi Jyothi} {and} \bibinfo{person}{Mark
  Hasegawa-Johnson}.} \bibinfo{year}{2015}\natexlab{}.
\newblock \showarticletitle{Improved Hindi broadcast ASR by adapting the
  language model and pronunciation model using a priori syntactic and
  morphophonemic knowledge}. In \bibinfo{booktitle}{\emph{Sixteenth Annual
  Conference of the International Speech Communication Association}}.
\newblock


\bibitem[\protect\citeauthoryear{Kitaev and Klein}{Kitaev and Klein}{2018}]%
        {kitaev2018constituency}
\bibfield{author}{\bibinfo{person}{Nikita Kitaev} {and} \bibinfo{person}{Dan
  Klein}.} \bibinfo{year}{2018}\natexlab{}.
\newblock \showarticletitle{Constituency parsing with a self-attentive
  encoder}.
\newblock \bibinfo{journal}{\emph{arXiv preprint arXiv:1805.01052}}
  (\bibinfo{year}{2018}).
\newblock


\bibitem[\protect\citeauthoryear{Krizhevsky, Sutskever, and Hinton}{Krizhevsky
  et~al\mbox{.}}{2012}]%
        {krizhevsky2012imagenet}
\bibfield{author}{\bibinfo{person}{Alex Krizhevsky}, \bibinfo{person}{Ilya
  Sutskever}, {and} \bibinfo{person}{Geoffrey~E Hinton}.}
  \bibinfo{year}{2012}\natexlab{}.
\newblock \showarticletitle{Imagenet classification with deep convolutional
  neural networks}.
\newblock \bibinfo{journal}{\emph{Advances in neural information processing
  systems}}  \bibinfo{volume}{25} (\bibinfo{year}{2012}),
  \bibinfo{pages}{1097--1105}.
\newblock


\bibitem[\protect\citeauthoryear{Kumar, Aggarwal, Mahata, Shah, Kumaraguru, and
  Zimmermann}{Kumar et~al\mbox{.}}{2019}]%
        {kumar2019get}
\bibfield{author}{\bibinfo{person}{Yaman Kumar}, \bibinfo{person}{Swati
  Aggarwal}, \bibinfo{person}{Debanjan Mahata}, \bibinfo{person}{Rajiv~Ratn
  Shah}, \bibinfo{person}{Ponnurangam Kumaraguru}, {and} \bibinfo{person}{Roger
  Zimmermann}.} \bibinfo{year}{2019}\natexlab{}.
\newblock \showarticletitle{Get IT Scored Using AutoSAS—An Automated System
  for Scoring Short Answers}. In \bibinfo{booktitle}{\emph{Proceedings of the
  AAAI Conference on Artificial Intelligence}}, Vol.~\bibinfo{volume}{33}.
  \bibinfo{pages}{9662--9669}.
\newblock


\bibitem[\protect\citeauthoryear{Kyriakopoulos, Knill, and Gales}{Kyriakopoulos
  et~al\mbox{.}}{2020}]%
        {kyriakopoulos2020automatic}
\bibfield{author}{\bibinfo{person}{Konstantinos Kyriakopoulos},
  \bibinfo{person}{Katherine Knill}, {and} \bibinfo{person}{Mark Gales}.}
  \bibinfo{year}{2020}\natexlab{}.
\newblock \showarticletitle{Automatic detection of accent and lexical
  pronunciation errors in spontaneous non-native English speech}.
\newblock  (\bibinfo{year}{2020}).
\newblock


\bibitem[\protect\citeauthoryear{Lee and Kim}{Lee and Kim}{2019}]%
        {lee2019robust}
\bibfield{author}{\bibinfo{person}{Younggun Lee} {and} \bibinfo{person}{Taesu
  Kim}.} \bibinfo{year}{2019}\natexlab{}.
\newblock \showarticletitle{Robust and fine-grained prosody control of
  end-to-end speech synthesis}. In \bibinfo{booktitle}{\emph{ICASSP 2019-2019
  IEEE International Conference on Acoustics, Speech and Signal Processing
  (ICASSP)}}. IEEE, \bibinfo{pages}{5911--5915}.
\newblock


\bibitem[\protect\citeauthoryear{Liu, Yang, Chi, Hsu, and Lee}{Liu
  et~al\mbox{.}}{2020}]%
        {liu2020Mockingjay}
\bibfield{author}{\bibinfo{person}{Andy~T Liu}, \bibinfo{person}{Shu-wen Yang},
  \bibinfo{person}{Po-Han Chi}, \bibinfo{person}{Po-chun Hsu}, {and}
  \bibinfo{person}{Hung-yi Lee}.} \bibinfo{year}{2020}\natexlab{}.
\newblock \showarticletitle{Mockingjay: Unsupervised speech representation
  learning with deep bidirectional transformer encoders}. In
  \bibinfo{booktitle}{\emph{ICASSP 2020-2020 IEEE International Conference on
  Acoustics, Speech and Signal Processing (ICASSP)}}. IEEE,
  \bibinfo{pages}{6419--6423}.
\newblock


\bibitem[\protect\citeauthoryear{McFee, Raffel, Liang, Ellis, McVicar,
  Battenberg, and Nieto}{McFee et~al\mbox{.}}{2015}]%
        {mcfee2015librosa}
\bibfield{author}{\bibinfo{person}{Brian McFee}, \bibinfo{person}{Colin
  Raffel}, \bibinfo{person}{Dawen Liang}, \bibinfo{person}{Daniel~PW Ellis},
  \bibinfo{person}{Matt McVicar}, \bibinfo{person}{Eric Battenberg}, {and}
  \bibinfo{person}{Oriol Nieto}.} \bibinfo{year}{2015}\natexlab{}.
\newblock \showarticletitle{librosa: Audio and music signal analysis in
  python}. In \bibinfo{booktitle}{\emph{Proceedings of the 14th python in
  science conference}}, Vol.~\bibinfo{volume}{8}. Citeseer,
  \bibinfo{pages}{18--25}.
\newblock


\bibitem[\protect\citeauthoryear{M{\"o}hle}{M{\"o}hle}{1984}]%
        {mohle1984comparison}
\bibfield{author}{\bibinfo{person}{Dorothea M{\"o}hle}.}
  \bibinfo{year}{1984}\natexlab{}.
\newblock \showarticletitle{A comparison of the second language speech
  production of different native speakers}.
\newblock \bibinfo{journal}{\emph{Second language productions}}
  \bibinfo{volume}{26} (\bibinfo{year}{1984}), \bibinfo{pages}{49}.
\newblock


\bibitem[\protect\citeauthoryear{Mooney}{Mooney}{2014}]%
        {mooneyQuip}
\bibfield{author}{\bibinfo{person}{Ray Mooney}.}
  \bibinfo{year}{2014}\natexlab{}.
\newblock \bibinfo{title}{You can't cram the meaning of a whole \%\&!\$\#
  sentence into a single \$\&!\#* vector!}
\newblock
  \bibinfo{howpublished}{\url{https://www.cs.utexas.edu/~mooney/cramming.html}}.
\newblock


\bibitem[\protect\citeauthoryear{Mulholland, Lopez, Evanini, Loukina, and
  Qian}{Mulholland et~al\mbox{.}}{2016}]%
        {mulholland2016comparison}
\bibfield{author}{\bibinfo{person}{Matthew Mulholland},
  \bibinfo{person}{Melissa Lopez}, \bibinfo{person}{Keelan Evanini},
  \bibinfo{person}{Anastassia Loukina}, {and} \bibinfo{person}{Yao Qian}.}
  \bibinfo{year}{2016}\natexlab{}.
\newblock \showarticletitle{A comparison of ASR and human errors for
  transcription of non-native spontaneous speech}. In
  \bibinfo{booktitle}{\emph{2016 IEEE International Conference on Acoustics,
  Speech and Signal Processing (ICASSP)}}. IEEE, \bibinfo{pages}{5855--5859}.
\newblock


\bibitem[\protect\citeauthoryear{Nagamine, Seltzer, and Mesgarani}{Nagamine
  et~al\mbox{.}}{2015}]%
        {nagamine2015exploring}
\bibfield{author}{\bibinfo{person}{Tasha Nagamine}, \bibinfo{person}{Michael~L
  Seltzer}, {and} \bibinfo{person}{Nima Mesgarani}.}
  \bibinfo{year}{2015}\natexlab{}.
\newblock \showarticletitle{Exploring how deep neural networks form phonemic
  categories}. In \bibinfo{booktitle}{\emph{Sixteenth Annual Conference of the
  International Speech Communication Association}}.
\newblock


\bibitem[\protect\citeauthoryear{Nagamine, Seltzer, and Mesgarani}{Nagamine
  et~al\mbox{.}}{2016}]%
        {nagamine2016role}
\bibfield{author}{\bibinfo{person}{Tasha Nagamine}, \bibinfo{person}{Michael~L
  Seltzer}, {and} \bibinfo{person}{Nima Mesgarani}.}
  \bibinfo{year}{2016}\natexlab{}.
\newblock \showarticletitle{On the Role of Nonlinear Transformations in Deep
  Neural Network Acoustic Models.}. In \bibinfo{booktitle}{\emph{Interspeech}}.
  \bibinfo{pages}{803--807}.
\newblock


\bibitem[\protect\citeauthoryear{Nagrani, Chung, and Zisserman}{Nagrani
  et~al\mbox{.}}{2017}]%
        {nagrani2017voxceleb}
\bibfield{author}{\bibinfo{person}{Arsha Nagrani}, \bibinfo{person}{Joon~Son
  Chung}, {and} \bibinfo{person}{Andrew Zisserman}.}
  \bibinfo{year}{2017}\natexlab{}.
\newblock \showarticletitle{Voxceleb: a large-scale speaker identification
  dataset}.
\newblock \bibinfo{journal}{\emph{arXiv preprint arXiv:1706.08612}}
  (\bibinfo{year}{2017}).
\newblock


\bibitem[\protect\citeauthoryear{Neumayer and Rauber}{Neumayer and
  Rauber}{2007}]%
        {neumayer2007integration}
\bibfield{author}{\bibinfo{person}{Robert Neumayer} {and}
  \bibinfo{person}{Andreas Rauber}.} \bibinfo{year}{2007}\natexlab{}.
\newblock \showarticletitle{Integration of text and audio features for genre
  classification in music information retrieval}. In
  \bibinfo{booktitle}{\emph{European Conference on Information Retrieval}}.
  Springer, \bibinfo{pages}{724--727}.
\newblock


\bibitem[\protect\citeauthoryear{Panayotov, Chen, Povey, and
  Khudanpur}{Panayotov et~al\mbox{.}}{2015}]%
        {panayotov2015librispeech}
\bibfield{author}{\bibinfo{person}{Vassil Panayotov}, \bibinfo{person}{Guoguo
  Chen}, \bibinfo{person}{Daniel Povey}, {and} \bibinfo{person}{Sanjeev
  Khudanpur}.} \bibinfo{year}{2015}\natexlab{}.
\newblock \showarticletitle{Librispeech: an asr corpus based on public domain
  audio books}. In \bibinfo{booktitle}{\emph{2015 IEEE International Conference
  on Acoustics, Speech and Signal Processing (ICASSP)}}. IEEE,
  \bibinfo{pages}{5206--5210}.
\newblock


\bibitem[\protect\citeauthoryear{Patil, Singla, Shah, Hama, and
  Zimmermann}{Patil et~al\mbox{.}}{2020}]%
        {patil2020towards}
\bibfield{author}{\bibinfo{person}{Rajaswa Patil}, \bibinfo{person}{Yaman~Kumar
  Singla}, \bibinfo{person}{Rajiv~Ratn Shah}, \bibinfo{person}{Mika Hama},
  {and} \bibinfo{person}{Roger Zimmermann}.} \bibinfo{year}{2020}\natexlab{}.
\newblock \showarticletitle{Towards Modelling Coherence in Spoken Discourse}.
\newblock \bibinfo{journal}{\emph{arXiv preprint arXiv:2101.00056}}
  (\bibinfo{year}{2020}).
\newblock


\bibitem[\protect\citeauthoryear{Peters, Neumann, Iyyer, Gardner, Clark, Lee,
  and Zettlemoyer}{Peters et~al\mbox{.}}{2018}]%
        {peters2018deep}
\bibfield{author}{\bibinfo{person}{Matthew~E Peters}, \bibinfo{person}{Mark
  Neumann}, \bibinfo{person}{Mohit Iyyer}, \bibinfo{person}{Matt Gardner},
  \bibinfo{person}{Christopher Clark}, \bibinfo{person}{Kenton Lee}, {and}
  \bibinfo{person}{Luke Zettlemoyer}.} \bibinfo{year}{2018}\natexlab{}.
\newblock \showarticletitle{Deep contextualized word representations}.
\newblock \bibinfo{journal}{\emph{arXiv preprint arXiv:1802.05365}}
  (\bibinfo{year}{2018}).
\newblock


\bibitem[\protect\citeauthoryear{Prasad and Jyothi}{Prasad and Jyothi}{2020}]%
        {prasad2020accents}
\bibfield{author}{\bibinfo{person}{Archiki Prasad} {and}
  \bibinfo{person}{Preethi Jyothi}.} \bibinfo{year}{2020}\natexlab{}.
\newblock \showarticletitle{How Accents Confound: Probing for Accent
  Information in End-to-End Speech Recognition Systems}. In
  \bibinfo{booktitle}{\emph{Proceedings of the 58th Annual Meeting of the
  Association for Computational Linguistics}}. \bibinfo{pages}{3739--3753}.
\newblock


\bibitem[\protect\citeauthoryear{Raj, Snyder, Povey, and Khudanpur}{Raj
  et~al\mbox{.}}{2019}]%
        {raj2019probing}
\bibfield{author}{\bibinfo{person}{Desh Raj}, \bibinfo{person}{David Snyder},
  \bibinfo{person}{Daniel Povey}, {and} \bibinfo{person}{Sanjeev Khudanpur}.}
  \bibinfo{year}{2019}\natexlab{}.
\newblock \showarticletitle{Probing the information encoded in x-vectors}. In
  \bibinfo{booktitle}{\emph{2019 IEEE Automatic Speech Recognition and
  Understanding Workshop (ASRU)}}. IEEE, \bibinfo{pages}{726--733}.
\newblock


\bibitem[\protect\citeauthoryear{Ramnath, Nema, Sahni, and Khapra}{Ramnath
  et~al\mbox{.}}{2020}]%
        {ramnath2020towards}
\bibfield{author}{\bibinfo{person}{Sahana Ramnath}, \bibinfo{person}{Preksha
  Nema}, \bibinfo{person}{Deep Sahni}, {and} \bibinfo{person}{Mitesh~M
  Khapra}.} \bibinfo{year}{2020}\natexlab{}.
\newblock \showarticletitle{Towards Interpreting BERT for Reading Comprehension
  Based QA}.
\newblock \bibinfo{journal}{\emph{arXiv preprint arXiv:2010.08983}}
  (\bibinfo{year}{2020}).
\newblock


\bibitem[\protect\citeauthoryear{Rasinski}{Rasinski}{2004}]%
        {rasinski2004assessing}
\bibfield{author}{\bibinfo{person}{Timothy~V Rasinski}.}
  \bibinfo{year}{2004}\natexlab{}.
\newblock \showarticletitle{Assessing reading fluency.}
\newblock \bibinfo{journal}{\emph{Pacific Resources for Education and Learning
  (PREL)}} (\bibinfo{year}{2004}).
\newblock


\bibitem[\protect\citeauthoryear{Read, Nation, et~al\mbox{.}}{Read
  et~al\mbox{.}}{2006}]%
        {read2006investigation}
\bibfield{author}{\bibinfo{person}{John Read}, \bibinfo{person}{Paul Nation},
  {et~al\mbox{.}}} \bibinfo{year}{2006}\natexlab{}.
\newblock \showarticletitle{An investigation of the lexical dimension of the
  IELTS speaking test}.
\newblock \bibinfo{journal}{\emph{IELTS research reports}}  \bibinfo{volume}{6}
  (\bibinfo{year}{2006}), \bibinfo{pages}{207--231}.
\newblock


\bibitem[\protect\citeauthoryear{Simonetta, Ntalampiras, and
  Avanzini}{Simonetta et~al\mbox{.}}{2019}]%
        {simonetta2019multimodal}
\bibfield{author}{\bibinfo{person}{Federico Simonetta},
  \bibinfo{person}{Stavros Ntalampiras}, {and} \bibinfo{person}{Federico
  Avanzini}.} \bibinfo{year}{2019}\natexlab{}.
\newblock \showarticletitle{Multimodal music information processing and
  retrieval: Survey and future challenges}. In \bibinfo{booktitle}{\emph{2019
  International Workshop on Multilayer Music Representation and Processing
  (MMRP)}}. IEEE, \bibinfo{pages}{10--18}.
\newblock


\bibitem[\protect\citeauthoryear{Tang, Ji, Li, and Zhou}{Tang
  et~al\mbox{.}}{2020}]%
        {tang2020dependency}
\bibfield{author}{\bibinfo{person}{Hao Tang}, \bibinfo{person}{Donghong Ji},
  \bibinfo{person}{Chenliang Li}, {and} \bibinfo{person}{Qiji Zhou}.}
  \bibinfo{year}{2020}\natexlab{}.
\newblock \showarticletitle{Dependency graph enhanced dual-transformer
  structure for aspect-based sentiment classification}. In
  \bibinfo{booktitle}{\emph{Proceedings of the 58th Annual Meeting of the
  Association for Computational Linguistics}}. \bibinfo{pages}{6578--6588}.
\newblock


\bibitem[\protect\citeauthoryear{Tian, Yi, Bai, Tao, Zhang, and Wen}{Tian
  et~al\mbox{.}}{2020}]%
        {tian2020synchronous}
\bibfield{author}{\bibinfo{person}{Zhengkun Tian}, \bibinfo{person}{Jiangyan
  Yi}, \bibinfo{person}{Ye Bai}, \bibinfo{person}{Jianhua Tao},
  \bibinfo{person}{Shuai Zhang}, {and} \bibinfo{person}{Zhengqi Wen}.}
  \bibinfo{year}{2020}\natexlab{}.
\newblock \showarticletitle{Synchronous transformers for end-to-end speech
  recognition}. In \bibinfo{booktitle}{\emph{ICASSP 2020-2020 IEEE
  International Conference on Acoustics, Speech and Signal Processing
  (ICASSP)}}. IEEE, \bibinfo{pages}{7884--7888}.
\newblock


\bibitem[\protect\citeauthoryear{Vaswani, Shazeer, Parmar, Uszkoreit, Jones,
  Gomez, Kaiser, and Polosukhin}{Vaswani et~al\mbox{.}}{2017}]%
        {vaswani2017attention}
\bibfield{author}{\bibinfo{person}{Ashish Vaswani}, \bibinfo{person}{Noam
  Shazeer}, \bibinfo{person}{Niki Parmar}, \bibinfo{person}{Jakob Uszkoreit},
  \bibinfo{person}{Llion Jones}, \bibinfo{person}{Aidan~N Gomez},
  \bibinfo{person}{{\L}ukasz Kaiser}, {and} \bibinfo{person}{Illia
  Polosukhin}.} \bibinfo{year}{2017}\natexlab{}.
\newblock \showarticletitle{Attention is all you need}. In
  \bibinfo{booktitle}{\emph{Advances in neural information processing
  systems}}. \bibinfo{pages}{5998--6008}.
\newblock


\bibitem[\protect\citeauthoryear{Wood}{Wood}{2001}]%
        {wood2001search}
\bibfield{author}{\bibinfo{person}{David Wood}.}
  \bibinfo{year}{2001}\natexlab{}.
\newblock \showarticletitle{In search of fluency: What is it and how can we
  teach it?}
\newblock \bibinfo{journal}{\emph{Canadian Modern Language Review}}
  \bibinfo{volume}{57}, \bibinfo{number}{4} (\bibinfo{year}{2001}),
  \bibinfo{pages}{573--589}.
\newblock


\bibitem[\protect\citeauthoryear{Wu, Wang, Pino, and Gu}{Wu
  et~al\mbox{.}}{2020}]%
        {wu2020self}
\bibfield{author}{\bibinfo{person}{Anne Wu}, \bibinfo{person}{Changhan Wang},
  \bibinfo{person}{Juan Pino}, {and} \bibinfo{person}{Jiatao Gu}.}
  \bibinfo{year}{2020}\natexlab{}.
\newblock \showarticletitle{Self-supervised representations improve end-to-end
  speech translation}.
\newblock \bibinfo{journal}{\emph{arXiv preprint arXiv:2006.12124}}
  (\bibinfo{year}{2020}).
\newblock


\bibitem[\protect\citeauthoryear{Wu and King}{Wu and King}{2016}]%
        {wu2016investigating}
\bibfield{author}{\bibinfo{person}{Zhizheng Wu} {and} \bibinfo{person}{Simon
  King}.} \bibinfo{year}{2016}\natexlab{}.
\newblock \showarticletitle{Investigating gated recurrent networks for speech
  synthesis}. In \bibinfo{booktitle}{\emph{2016 IEEE International Conference
  on Acoustics, Speech and Signal Processing (ICASSP)}}. IEEE,
  \bibinfo{pages}{5140--5144}.
\newblock


\bibitem[\protect\citeauthoryear{Yan, Kim, and Kim}{Yan et~al\mbox{.}}{2018}]%
        {yan2018complexity}
\bibfield{author}{\bibinfo{person}{Xun Yan}, \bibinfo{person}{Ha~Ram Kim},
  {and} \bibinfo{person}{J Kim}.} \bibinfo{year}{2018}\natexlab{}.
\newblock \bibinfo{title}{Complexity, Accuracy and Fluency (CAF). Features of
  speaking performances on APTIS across different levels on the common European
  framework of reference for languages (CEFR)}.
\newblock
\newblock


\bibitem[\protect\citeauthoryear{Zhang, Ramachandran, Tenney, Elazar, and
  Roth}{Zhang et~al\mbox{.}}{2020}]%
        {zhang2020language}
\bibfield{author}{\bibinfo{person}{Xikun Zhang}, \bibinfo{person}{Deepak
  Ramachandran}, \bibinfo{person}{Ian Tenney}, \bibinfo{person}{Yanai Elazar},
  {and} \bibinfo{person}{Dan Roth}.} \bibinfo{year}{2020}\natexlab{}.
\newblock \showarticletitle{Do Language Embeddings Capture Scales?}
\newblock \bibinfo{journal}{\emph{arXiv preprint arXiv:2010.05345}}
  (\bibinfo{year}{2020}).
\newblock


\bibitem[\protect\citeauthoryear{Zhang and Duan}{Zhang and Duan}{2018}]%
        {zhang2018visualization}
\bibfield{author}{\bibinfo{person}{Yichi Zhang} {and} \bibinfo{person}{Zhiyao
  Duan}.} \bibinfo{year}{2018}\natexlab{}.
\newblock \showarticletitle{Visualization and interpretation of Siamese style
  convolutional neural networks for sound search by vocal imitation}. In
  \bibinfo{booktitle}{\emph{2018 IEEE International Conference on Acoustics,
  Speech and Signal Processing (ICASSP)}}. IEEE, \bibinfo{pages}{2406--2410}.
\newblock


\bibitem[\protect\citeauthoryear{Zhang, Weiss, Zen, Wu, Chen, Skerry-Ryan, Jia,
  Rosenberg, and Ramabhadran}{Zhang et~al\mbox{.}}{2019}]%
        {zhang2019learning}
\bibfield{author}{\bibinfo{person}{Yu Zhang}, \bibinfo{person}{Ron~J Weiss},
  \bibinfo{person}{Heiga Zen}, \bibinfo{person}{Yonghui Wu},
  \bibinfo{person}{Zhifeng Chen}, \bibinfo{person}{RJ Skerry-Ryan},
  \bibinfo{person}{Ye Jia}, \bibinfo{person}{Andrew Rosenberg}, {and}
  \bibinfo{person}{Bhuvana Ramabhadran}.} \bibinfo{year}{2019}\natexlab{}.
\newblock \showarticletitle{Learning to speak fluently in a foreign language:
  Multilingual speech synthesis and cross-language voice cloning}.
\newblock \bibinfo{journal}{\emph{arXiv preprint arXiv:1907.04448}}
  (\bibinfo{year}{2019}).
\newblock


\bibitem[\protect\citeauthoryear{{Zhao}, {Sonsaat}, {Silpachai}, {Lucic},
  {Chukharev-Hudilainen}, {Levis}, and {Gutierrez-Osuna}}{{Zhao}
  et~al\mbox{.}}{2018}]%
        {zhao2018l2arctic}
\bibfield{author}{\bibinfo{person}{Guanlong {Zhao}}, \bibinfo{person}{Sinem
  {Sonsaat}}, \bibinfo{person}{Alif {Silpachai}}, \bibinfo{person}{Ivana
  {Lucic}}, \bibinfo{person}{Evgeny {Chukharev-Hudilainen}},
  \bibinfo{person}{John {Levis}}, {and} \bibinfo{person}{Ricardo
  {Gutierrez-Osuna}}.} \bibinfo{year}{2018}\natexlab{}.
\newblock \showarticletitle{L2-ARCTIC: A Non-native English Speech Corpus}. In
  \bibinfo{booktitle}{\emph{Proc. Interspeech}}. \bibinfo{pages}{2783–2787}.
\newblock
\urldef\tempurl%
\url{https://doi.org/10.21437/Interspeech.2018-1110}
\showDOI{\tempurl}


\end{thebibliography}

\appendix

\section{Experimental results}
\begin{table*}[!htbp]
\begin{adjustbox}{width=1\textwidth}
\begin{tabular}{@{}l|rrrrrrrrrrrr@{}}
\toprule
\textbf{Features\textbackslash  Layers} & 1 & 2 & 3 & 4 & 5 & 6 & 7 & 8 & 9 & 10 & 11 & 12 \\ \midrule
Duration & 0.001001 & 0.001053 & 0.001099 & 0.000988 & \textbf{0.000893} & 0.001037 & 0.000985 & 0.001205 & 0.001173 & 0.001198 & 0.001597 & 0.001742 \\
stdev\_energy & \textbf{0.002307} & 0.00238 & 0.002714 & 0.003584 & 0.004343 & 0.004559 & 0.004829 & 0.004668 & 0.004243 & 0.003856 & 0.004065 & 0.004143 \\
mean\_pitch & \textbf{0.001812} & 0.001821 & 0.002083 & 0.00279 & 0.003374 & 0.003993 & 0.003758 & 0.00421 & 0.003715 & 0.002609 & 0.003511 & 0.003896 \\
voiced\_to\_unvoiced\_ratio & 0.002232 & 0.002014 & 0.001971 & 0.002092 & 0.00225 & 0.002457 & 0.00244 & 0.002375 & 0.00236 & \textbf{0.001863} & 0.002477 & 0.002677 \\
zero\_crossing\_rate & 0.0044 & \textbf{0.004152} & 0.004536 & 0.005638 & 0.006737 & 0.007006 & 0.008681 & 0.007473 & 0.006309 & 0.006546 & 0.007058 & 0.007533 \\
energy\_entropy & 0.004003 & \textbf{0.003852} & 0.004065 & 0.004144 & 0.004913 & 0.004166 & 0.004566 & 0.00414 & 0.0042 & 0.004021 & 0.004928 & 0.005935 \\
spectral\_centroid & 0.000335 & \textbf{0.000335} & 0.000335 & 0.000335 & 0.000335 & 0.000335 & 0.000335 & 0.000335 & 0.000335 & 0.000335 & 0.000335 & 0.000335 \\
localJitter & 0.002261 & \textbf{0.001933} & 0.002131 & 0.002111 & 0.002273 & 0.002446 & 0.002638 & 0.002834 & 0.002379 & 0.001996 & 0.002267 & 0.002489 \\
localShimmer & \textbf{0.003059} & 0.003057 & 0.003449 & 0.003517 & 0.004047 & 0.004389 & 0.004901 & 0.004478 & 0.004077 & 0.00355 & 0.004037 & 0.00396 \\ \bottomrule
\end{tabular}
\end{adjustbox}
\vspace{1 mm}
\caption{\label{Audio_W}  \small Results (MSE) for audio features on wav2vec2.0 for native read speech corpus (Librispeech) %\cy{Font of this and the below tables are too small, you should split it into multiple rows.}
}
\vspace{-3 mm}
\end{table*}

\begin{table*}[!htbp]
\begin{adjustbox}{width=1\textwidth}
\begin{tabular}{@{}l|rrrrrrrrrrrr@{}}
\toprule
\textbf{Features\textbackslash Layers} & 1 & 2 & 3 & 4 & 5 & 6 & 7 & 8 & 9 & 10 & 11 & 12 \\ \midrule
filled\_pause\_rate & 0.000877 & 0.00093 & 0.00083 & 0.000917 & 0.000831 & 0.000838 & 0.00081 & \textbf{0.000782} & 0.000794 & 0.000802 & 0.000815 & 0.000829 \\
general\_silence & 0.001896 & 0.001805 & 0.00194 & 0.001795 & \textbf{0.001684} & 0.001924 & 0.002031 & 0.001937 & 0.00198 & 0.002098 & 0.002722 & 0.002112 \\
mean\_silence & 0.001807 & 0.001908 & 0.001891 & 0.001959 & 0.001821 & \textbf{0.001723} & 0.001787 & 0.001845 & 0.001906 & 0.001886 & 0.002394 & 0.002328 \\
silence\_abs\_deviation & \textbf{0.000975} & 0.001266 & 0.00149 & 0.001221 & 0.001371 & 0.001316 & 0.00158 & 0.001618 & 0.001493 & 0.001484 & 0.001869 & 0.001599 \\
SilenceRate1 & 0.005096 & 0.004997 & 0.005074 & 0.004758 & 0.004217 & \textbf{0.003676} & 0.003839 & 0.004035 & 0.004023 & 0.00436 & 0.004927 & 0.005627 \\
SilenceRate2 & 0.00516 & 0.00552 & 0.005248 & 0.004933 & 0.005085 & 0.004941 & \textbf{0.004845} & 0.005112 & 0.005174 & 0.00574 & 0.005895 & 0.006605 \\
speaking\_rate & 0.013043 & 0.012784 & 0.012493 & 0.010239 & 0.007733 & 0.006184 & 0.005216 & \textbf{0.005029} & 0.005487 & 0.006623 & 0.009679 & 0.01164 \\
articulation\_rate & 0.016824 & 0.015866 & 0.014793 & 0.012394 & 0.008917 & 0.007374 & 0.006135 & 0.006321 & \textbf{0.006001} & 0.007958 & 0.010589 & 0.011723 \\
longpfreq & \textbf{0.001642} & 0.001979 & 0.001774 & 0.001982 & 0.001731 & 0.001646 & 0.001798 & 0.001868 & 0.001698 & 0.001848 & 0.001863 & 0.001995 \\
average\_syllables\_in\_words & 0.018313 & 0.018215 & 0.018016 & 0.015562 & 0.01167 & 0.008615 & 0.006652 & \textbf{0.006486} & 0.007123 & 0.010458 & 0.013834 & 0.014869 \\
wordsyll2 & 0.010109 & 0.010726 & 0.011357 & 0.009438 & 0.007506 & 0.006293 & 0.005058 & \textbf{0.005559} & 0.005261 & 0.005988 & 0.008059 & 0.007614 \\
repetition\_freq & 0.015586 & 0.016036 & 0.017251 & 0.016137 & 0.015412 & 0.014447 & 0.013344 & 0.01352 & \textbf{0.013318} & 0.015553 & 0.014703 & 0.013719
\\\bottomrule
\end{tabular}
\end{adjustbox}
\vspace{1 mm}
\caption{\label{Fluency_W} \small  Results (MSE) for fluency features on wav2vec2.0 for native read speech corpus (Librispeech)}

\end{table*}

\begin{table*}[!htb]
\begin{adjustbox}{width=1\textwidth}
\begin{tabular}{@{}l|rrrrrrrrrrrr@{}}
\toprule
\textbf{Features\textbackslash{}Layers} & 1 & 2 & 3 & 4 & 5 & 6 & 7 & 8 & 9 & 10 & 11 & 12 \\ \midrule
StressedSyllPercent & 0.018905 & 0.019561 & 0.020112 & 0.018046 & 0.020365 & 0.002052 & 0.001923 & 0.001946 & \textbf{0.001741} & 0.002095 & 0.002429 & 0.002547 \\
StressDistanceSyllMean & 0.00951 & 0.009831 & 0.009026 & 0.00883 & 0.008503 & 0.008189 & 0.007759 & 0.007837 & 0.008494 & \textbf{0.00752} & 0.00922 & 0.008768 \\
StressDistanceMean & 0.01204 & 0.012537 & 0.012932 & 0.013452 & \textbf{0.01043} & 0.010614 & 0.011099 & 0.010778 & 0.010512 & 0.011903 & 0.011996 & 0.011983 \\
vowelPercentage & 0.007322 & 0.007084 & 0.006278 & 0.005989 & 0.005836 & 0.005385 & 0.005205 & 0.005008 & \textbf{0.004806} & 0.00545 & 0.006394 & 0.006526 \\
consonantPercentage & 0.00597 & 0.006529 & 0.007062 & 0.005323 & 0.005464 & 0.004765 & 0.004961 & 0.005012 & 0.004824 & \textbf{0.004472} & 0.005979 & 0.006222 \\
vowelDurationSD & 0.002792 & 0.002813 & 0.002516 & 0.002353 & 0.002159 & 0.002076 & 0.001875 & 0.001894 & \textbf{0.001866} & 0.001997 & 0.002379 & 0.002423 \\
consonantDurationSD & 0.001645 & 0.001361 & 0.001378 & 0.001352 & 0.001319 & 0.001274 & \textbf{0.001254} & 0.001392 & 0.001346 & 0.001282 & 0.001411 & 0.00154 \\
syllableDurationSD & 0.005843 & 0.005444 & 0.005413 & 0.004813 & 0.004507 & 0.003987 & \textbf{0.003908} & 0.003954 & 0.004033 & 0.004483 & 0.005114 & 0.005243 \\
vowelSDNorm & 0.003604 & 0.003876 & 0.003826 & 0.003511 & 0.003458 & 0.003277 & 0.003311 & 0.003346 & \textbf{0.00323} & 0.003449 & 0.003971 & 0.003586 \\
consonantSDNorm & 0.002454 & 0.002599 & 0.002593 & 0.002414 & 0.002399 & 0.0023 & \textbf{0.002272} & 0.002299 & 0.002389 & 0.002296 & 0.00252 & 0.00247 \\
syllableSDNorm & 0.006879 & 0.006955 & 0.007364 & 0.006514 & 0.00589 & 0.005689 & \textbf{0.005543} & 0.005619 & 0.005713 & 0.006977 & 0.007221 & 0.007269 \\
vowelPVINorm & 0.008083 & 0.008477 & 0.008793 & 0.00854 & \textbf{0.007294} & 0.007872 & 0.00736 & 0.008043 & 0.007457 & 0.008197 & 0.008581 & 0.008054 \\
consonantPVINorm & 0.007041 & 0.007458 & 0.007528 & 0.006909 & \textbf{0.006419} & 0.006852 & 0.006777 & 0.006805 & 0.007341 & 0.006959 & 0.007225 & 0.007434 \\
syllablePVINorm & 0.012702 & 0.013669 & 0.013096 & 0.012062 & 0.012077 & 0.011151 & \textbf{0.011088} & 0.011425 & 0.011091 & 0.01258 & 0.013119 & 0.013045 \\ \bottomrule
\end{tabular}
\end{adjustbox}
\vspace{1 mm}
\caption{\label{Pron_W} \small  Results (MSE) for pronunciation features on wav2vec2.0 for native read speech corpus (Librispeech)}

\end{table*}

\begin{table*}[]
\begin{adjustbox}{width=1\textwidth}
\begin{tabular}{@{}l|rrrrrrrrrrrr@{}}
\toprule
\textbf{Features\textbackslash{}Layers} & 1        & 2        & 3        & 4        & 5        & 6        & 7        & 8        & 9        & 10       & 11       & 12       \\ \midrule
Unique Word count              & 0.005403 & 0.004279 & 0.006434 & 0.007045 & \textbf{0.004074} & 0.005392          & 0.004464          & 0.004928          & 0.007097          & 0.005556 & 0.005561 & 0.005696          \\
Word Complexity                & 0.011001 & 0.010729 & 0.010597 & 0.010257 & 0.009527          & 0.009294          & \textbf{0.00883}  & 0.009122          & 0.009118          & 0.009888 & 0.010307 & 0.010515          \\
Total adjectives               & 0.008716 & 0.008636 & 0.008345 & 0.009197 & 0.009202          & 0.009062          & \textbf{0.007672} & 0.008421          & 0.007952          & 0.009323 & 0.009411 & 0.009366          \\
Total adverbs                  & 0.011648 & 0.011361 & 0.01115  & 0.010896 & 0.009901          & 0.00901           & \textbf{0.008756} & 0.00871           & 0.009569          & 0.010773 & 0.010703 & 0.01076           \\
Total nouns                    & 0.004879 & 0.005831 & 0.005286 & 0.004014 & 0.004326          & 0.004439          & 0.004539          & 0.004013          & \textbf{0.00385}  & 0.004392 & 0.004874 & 0.005444          \\
Total verbs                    & 0.009748 & 0.008873 & 0.009065 & 0.008777 & 0.007376          & \textbf{0.006792} & 0.00808           & 0.007944          & 0.007044          & 0.009047 & 0.009181 & 0.008955          \\
Total pronoun                  & 0.002278 & 0.002251 & 0.00228  & 0.002274 & 0.002318          & 0.0021            & 0.002364          & 0.002101          & \textbf{0.001889} & 0.002211 & 0.00224  & 0.002304          \\
Total conjunction              & 0.004891 & 0.004882 & 0.005034 & 0.004929 & 0.004822          & 0.004451          & 0.004238          & \textbf{0.004201} & 0.004505          & 0.00487  & 0.004885 & 0.004949          \\
Total determiners              & 0.001954 & 0.001966 & 0.001957 & 0.00193  & 0.001954          & 0.002219          & 0.00232           & \textbf{0.001854} & 0.001931          & 0.001946 & 0.001956 & 0.001953          \\
No. of subj                    & 0.014399 & 0.016879 & 0.01675  & 0.016729 & 0.016072          & 0.014606          & \textbf{0.01289}  & 0.015839          & 0.013978          & 0.013527 & 0.015682 & 0.013509          \\
No. of obj                     & 0.017454 & 0.01987  & 0.022306 & 0.021348 & 0.02231           & 0.022954          & 0.018353          & 0.020034          & 0.019796          & 0.018132 & 0.017848 & \textbf{0.016293} \\
Tree depth                     & 0.010957 & 0.01198  & 0.013783 & 0.01487  & 0.013179          & 0.017128          & 0.011726          & 0.016646          & 0.012092          & 0.011978 & 0.012764 & \textbf{0.009952}\\ \bottomrule
\end{tabular}
\end{adjustbox}
\vspace{1 mm}
\caption{\label{Vocab_M} \small  Results (MSE) for text features on {\wv} for native read speech corpus (Librispeech)}
\end{table*}

\begin{table*}[]
\begin{adjustbox}{width=1\textwidth}
\begin{tabular}{@{}l|rrrrrrrrrrrr@{}}
\toprule
\textbf{Features\textbackslash{}Layers} & 1 & 2 & 3 & 4 & 5 & 6 & 7 & 8 & 9 & 10 & 11 & 12 \\ \midrule
total\_duration & \textbf{0.0006} & 0.001195 & 0.002857 & 0.003293 & 0.003334 & 0.003413 & 0.003017 & 0.004304 & 0.003943 & 0.005397 & 0.005148 & 0.005897 \\
stdev\_energy & 0.007231 & 0.005821 & 0.008878 & 0.00692 & 0.006879 & 0.006747 & 0.006156 & 0.00652 & 0.005613 & 0.005844 & \textbf{0.005424} & 0.005705 \\
mean\_pitch & 0.002864 & 0.005313 & 0.009793 & 0.010587 & 0.009883 & 0.010169 & 0.009742 & 0.005038 & 0.003443 & 0.001832 & \textbf{0.000927} & 0.001272 \\
voiced\_to\_unvoiced\_ratio & 0.004341 & 0.003959 & 0.005517 & 0.005531 & 0.005119 & 0.005269 & 0.006042 & 0.004361 & 0.002816 & 0.002329 & 0.002053 & \textbf{0.001834} \\
zero\_crossing\_rate & 0.011892 & 0.013714 & 0.014845 & 0.015256 & 0.01248 & 0.013971 & 0.012568 & 0.013016 & 0.011436 & \textbf{0.007881} & 0.008748 & 0.009278 \\
energy\_entropy & 0.005736 & \textbf{0.005621} & 0.006332 & 0.006845 & 0.006357 & 0.006518 & 0.006331 & 0.006071 & 0.005919 & 0.00635 & 0.00684 & 0.00662 \\
spectral\_centroid & 0.000335 & 0.000335 & 0.000335 & 0.000335 & \textbf{0.000335} & 0.000335 & 0.000335 & 0.000336 & 0.000348 & 0.000335 & 0.000335 & 0.000335 \\
localJitter & 0.002802 & 0.002952 & 0.003129 & 0.003784 & 0.003684 & 0.003827 & 0.003192 & 0.0031 & 0.002831 & 0.002181 & 0.002076 & 0.001966 \\
localShimmer & 0.006434 & 0.006935 & 0.007138 & 0.007962 & 0.0072 & 0.007752 & 0.007866 & 0.007677 & 0.006455 & 0.005965 & 0.005424 & \textbf{0.005135}\\ \bottomrule
\end{tabular}
\end{adjustbox}
\vspace{1 mm}
\caption{\label{Audio_M}  \small Results (MSE) for audio features on {\mj} for native read speech corpus (Librispeech)}

\end{table*}

\begin{table*}[]
\begin{adjustbox}{width=1\textwidth}
\begin{tabular}{@{}l|rrrrrrrrrrrr@{}}
\toprule
\textbf{Features\textbackslash{}Layers} & 1 & 2 & 3 & 4 & 5 & 6 & 7 & 8 & 9 & 10 & 11 & 12 \\ \midrule
filled\_pause\_rate & 0.00079 & 0.000775 & 0.000776 & 0.000774 & 0.000773 & \textbf{0.000769} & 0.000775 & 0.00078 & 0.000785 & 0.000835 & 0.000815 & 0.000789 \\
general\_silence & 0.003559 & 0.003244 & 0.002682 & 0.003353 & 0.003067 & \textbf{0.002264} & 0.002346 & 0.002401 & 0.003544 & 0.004336 & 0.004376 & 0.0046 \\
mean\_silence & 0.002124 & 0.002661 & 0.002411 & 0.002158 & 0.001764 & 0.002073 & 0.001833 & 0.002313 & 0.001883 & 0.003345 & 0.001942 & \textbf{0.00174} \\
silence\_absolute\_deviation & 0.003095 & 0.002299 & 0.001743 & 0.001615 & 0.001543 & 0.001675 & \textbf{0.001372} & 0.002061 & 0.001954 & 0.00175 & 0.001509 & 0.002432 \\
SilenceRate1 & 0.005183 & 0.005171 & 0.005297 & 0.005822 & 0.005221 & 0.005045 & 0.005098 & 0.004941 & 0.004533 & 0.004755 & 0.004788 & \textbf{0.004162} \\
SilenceRate2 & 0.005451 & 0.004879 & 0.005217 & 0.005187 & 0.005731 & 0.005497 & 0.005417 & \textbf{0.004746} & 0.005169 & 0.005023 & 0.006062 & 0.006018 \\
speaking\_rate & 0.012803 & 0.012767 & 0.014038 & 0.014252 & 0.014247 & 0.014811 & 0.013957 & 0.015266 & 0.012086 & 0.01259 & \textbf{0.011059} & 0.011137 \\
articulation\_rate & 0.016709 & 0.016736 & 0.018223 & 0.018936 & 0.018751 & 0.018269 & 0.017996 & 0.019456 & 0.015512 & 0.014811 & \textbf{0.012795} & 0.013218 \\
longpfreq & \textbf{0.001566} & 0.001713 & 0.001707 & 0.001574 & 0.001661 & 0.001604 & 0.001672 & 0.001607 & 0.001571 & 0.001813 & 0.002025 & 0.0018 \\
average\_syllables\_in\_words & 0.014321 & \textbf{0.013737} & 0.01423 & 0.015298 & 0.015195 & 0.014701 & 0.01449 & 0.014178 & 0.014282 & 0.015107 & 0.015409 & 0.014309 \\
wordsyll2 & \textbf{0.010012} & 0.010246 & 0.012442 & 0.012758 & 0.011834 & 0.011637 & 0.011688 & 0.011879 & 0.011029 & 0.012438 & 0.012399 & 0.011999 \\
repetition\_freq & 0.011282 & \textbf{0.011277} & 0.011447 & 0.011672 & 0.011502 & 0.011495 & 0.011444 & 0.011428 & 0.011734 & 0.01204 & 0.012926 & 0.012278\\ \bottomrule
\end{tabular}
\end{adjustbox}
\vspace{1 mm}
\caption{\label{Fluency_M}  \small Results (MSE) for fluency features on {\mj} for native read speech corpus (Librispeech)}

\end{table*}

\begin{table*}[]
\begin{adjustbox}{width=1\textwidth}
\begin{tabular}{@{}l|rrrrrrrrrrrr@{}}
\toprule
\textbf{Features\textbackslash{}Layers} & 1 & 2 & 3 & 4 & 5 & 6 & 7 & 8 & 9 & 10 & 11 & 12 \\ \midrule
StressedSyllPercent & 0.016776 & 0.016777 & 0.017137 & 0.017021 & 0.01676 & 0.017471 & 0.016636 & 0.016884 & 0.016915 & \textbf{0.015961} & 0.016951 & 0.01643 \\
StressDistanceSyllMean & 0.002738 & 0.00502 & 0.008836 & 0.011745 & 0.009905 & 0.009778 & 0.009395 & 0.004853 & 0.003612 & 0.001711 & 0.00096 & \textbf{0.000799} \\
StressDistanceMean & 0.010448 & 0.011868 & 0.014704 & 0.016046 & 0.015573 & 0.015609 & 0.014304 & 0.011836 & 0.010375 & 0.007646 & 0.005007 & \textbf{0.004817} \\
vowelPercentage & 0.006115 & 0.006363 & 0.007021 & 0.007937 & 0.007496 & 0.007167 & 0.006866 & 0.005513 & 0.005456 & 0.004813 & \textbf{0.004648} & 0.004672 \\
consonantPercentage & 0.005199 & 0.005549 & 0.006474 & 0.00614 & 0.005955 & 0.005665 & 0.005455 & 0.005207 & 0.005086 & 0.004534 & \textbf{0.004377} & 0.004635 \\
vowelDurationSD & 0.002703 & 0.002683 & 0.002921 & 0.00286 & 0.003081 & 0.002907 & 0.002783 & 0.002688 & 0.002457 & 0.002194 & \textbf{0.002132} & 0.002288 \\
consonantDurationSD & 0.001128 & 0.001202 & 0.001373 & 0.001322 & 0.001312 & 0.001375 & 0.001264 & 0.001207 & 0.001089 & 0.001026 & 0.001074 & \textbf{0.000959} \\
syllableDurationSD & 0.004989 & 0.005503 & 0.005784 & 0.005818 & 0.005728 & 0.005975 & 0.005904 & 0.005596 & 0.004834 & 0.004513 & \textbf{0.004304} & 0.004305 \\
vowelSDNorm & 0.003146 & 0.002863 & 0.002959 & 0.002969 & 0.00301 & 0.002889 & 0.002947 & 0.002857 & 0.002926 & 0.002932 & 0.002937 & \textbf{0.00288} \\
consonantSDNorm & 0.001863 & 0.002026 & 0.001836 & 0.001915 & 0.001888 & 0.001876 & 0.001865 & 0.001871 & 0.001939 & 0.001928 & 0.001948 & \textbf{0.001851} \\
syllableSDNorm & 0.005986 & 0.005941 & 0.005994 & 0.005847 & 0.005853 & 0.005905 & 0.006004 & 0.005834 & 0.005789 & 0.005879 & 0.005828 & \textbf{0.005727} \\
vowelPVINorm & 0.006571 & 0.006466 & 0.006616 & 0.006877 & 0.006913 & 0.006888 & 0.006579 & 0.006727 & 0.006548 & 0.006447 & 0.006748 & \textbf{0.006356} \\
consonantPVINorm & 0.00558 & 0.005594 & 0.005786 & 0.005957 & 0.005809 & 0.005824 & 0.005664 & 0.005537 & 0.005373 & \textbf{0.005287} & 0.005391 & 0.005677 \\
syllablePVINorm & 0.010683 & 0.010746 & 0.011027 & 0.01068 & 0.010567 & 0.010839 & 0.0108 & 0.010717 & \textbf{0.010553} & 0.010659 & 0.011023 & 0.010602
\\ \bottomrule
\end{tabular}
\end{adjustbox}
\vspace{1 mm}
\caption{\label{Pron_M}  \small Results (MSE) for pronunciation features on {\mj} for native read speech corpus (Librispeech)}

\end{table*}

\begin{table*}[]
\begin{adjustbox}{width=1\textwidth}
\begin{tabular}{@{}l|rrrrrrrrrrrr@{}}
\toprule
\textbf{Features\textbackslash{}Layers} & 1                 & 2                 & 3                & 4        & 5        & 6        & 7        & 8        & 9        & 10       & 11       & 12       \\
Unique Word count                       & \textbf{0.002199} & 0.005996          & 0.00673          & 0.005919 & 0.006723 & 0.007532 & 0.005243 & 0.005413 & 0.006635 & 0.008469 & 0.011627 & 0.005578 \\
Word Complexity                         & \textbf{0.011128} & 0.011406          & 0.011535         & 0.011591 & 0.01167  & 0.011325 & 0.011527 & 0.01143  & 0.011467 & 0.012014 & 0.011469 & 0.011363 \\
Total adjectives                        & \textbf{0.00757}  & 0.00882           & 0.009862         & 0.011036 & 0.010323 & 0.011291 & 0.009432 & 0.009538 & 0.010743 & 0.011471 & 0.012513 & 0.010343 \\
Total adverbs                           & \textbf{0.010991} & 0.011332          & 0.011315         & 0.011537 & 0.011498 & 0.011458 & 0.011387 & 0.011352 & 0.011593 & 0.012657 & 0.013342 & 0.01247  \\
Total nouns                             & \textbf{0.004011} & 0.004433          & 0.006881         & 0.007137 & 0.006792 & 0.006114 & 0.005585 & 0.006673 & 0.007262 & 0.007768 & 0.008683 & 0.006561 \\
Total verbs                             & \textbf{0.008358} & 0.009059          & 0.010349         & 0.010684 & 0.010196 & 0.01044  & 0.009787 & 0.009147 & 0.012781 & 0.013148 & 0.014656 & 0.011548 \\
Total pronoun                           & 0.002238          & 0.002217          & \textbf{0.00221} & 0.002231 & 0.002242 & 0.002281 & 0.002264 & 0.002303 & 0.002259 & 0.002374 & 0.002525 & 0.002345 \\
Total conjunction                       & 0.004935          & \textbf{0.004902} & 0.004965         & 0.005017 & 0.005184 & 0.00505  & 0.004963 & 0.005009 & 0.005076 & 0.00517  & 0.00555  & 0.005081 \\
Total determiners                       & \textbf{0.00194}  & 0.001974          & 0.001944         & 0.001952 & 0.001968 & 0.001977 & 0.001941 & 0.001942 & 0.001977 & 0.002006 & 0.002023 & 0.001973 \\
No. of subj                             & 0.010997          & \textbf{0.010014} & 0.010586         & 0.011605 & 0.010631 & 0.011095 & 0.01074  & 0.010356 & 0.010707 & 0.011335 & 0.011679 & 0.011205 \\
No. of obj                              & 0.011757          & \textbf{0.011308} & 0.012495         & 0.012662 & 0.012242 & 0.012031 & 0.012256 & 0.012449 & 0.011985 & 0.01295  & 0.013274 & 0.013587 \\
Tree depth                              & \textbf{0.00611}  & 0.006107          & 0.007302         & 0.011605 & 0.008775 & 0.007191 & 0.007097 & 0.00715  & 0.006802 & 0.007231 & 0.007594 & 0.007813\\
\bottomrule
\end{tabular}
\end{adjustbox}
\vspace{1 mm}
\caption{\label{Vocab_M} \small  Results (MSE) for text features on {\mj} for native read speech corpus (Librispeech)}
\end{table*}

\begin{table*}[]
\begin{adjustbox}{width=1\textwidth}
\begin{tabular}{@{}l|rrrrrrrrrrrr@{}}
\toprule
\textbf{Features\textbackslash{}Layers} & 1 & 2 & 3 & 4 & 5 & 6 & 7 & 8 & 9 & 10 & 11 & 12 \\ \midrule
total\_duration & 0.003037 & \textbf{0.002241} & 0.002387 & 0.002949 & 0.002987 & 0.002997 & 0.002993 & 0.00398 & 0.003188 & 0.003082 & 0.00443 & 0.005753 \\
stdev\_energy & 0.013247 & \textbf{0.010778} & 0.013224 & 0.011689 & 0.011251 & 0.011181 & 0.011164 & 0.01123 & 0.012513 & 0.011796 & 0.011889 & 0.011455 \\
mean\_pitch & 0.004493 & \textbf{0.003569} & 0.003843 & 0.004897 & 0.005505 & 0.004684 & 0.005197 & 0.005699 & 0.005733 & 0.004194 & 0.008189 & 0.006506 \\
voiced\_to\_unvoiced\_ratio & 0.002074 & 0.002024 & 0.001661 & 0.002073 & 0.002288 & \textbf{0.001632} & 0.001988 & 0.001961 & 0.001982 & 0.001904 & 0.002233 & 0.002125 \\
zero\_crossing\_rate & 0.010519 & 0.007792 & 0.006901 & 0.007062 & 0.008208 & 0.00679 & 0.00718 & 0.006587 & 0.006429 & \textbf{0.006369} & 0.010559 & 0.009979 \\
energy\_entropy & 0.013166 & 0.010519 & \textbf{0.008692} & 0.01036 & 0.010414 & 0.01094 & 0.010786 & 0.010152 & 0.013525 & 0.009774 & 0.010588 & 0.011592 \\
spectral\_centroid & 0.000004 & 0.000003 & 0.000003 & 0.000003 & 0.000003 & 0.000003 & \textbf{0.000003} & 0.000004 & 0.000004 & 0.000003 & 0.000003 & 0.000003 \\
localJitter & 0.008843 & 0.00763 & 0.010089 & 0.007924 & \textbf{0.007034} & 0.008899 & 0.007446 & 0.01006 & 0.008025 & 0.007731 & 0.008643 & 0.008794 \\
localShimmer & 0.005666 & 0.00494 & 0.005273 & 0.004515 & 0.006485 & \textbf{0.004504} & 0.004659 & 0.005153 & 0.004648 & 0.005181 & 0.006095 & 0.005419
\\ \bottomrule
\end{tabular}
\end{adjustbox}
\vspace{1 mm}
\caption{\label{nAudio_W} \small  Results (MSE) for audio features on {\wv} for non-native read speech corpus (L2 Arctic)}

\end{table*}

\begin{table*}[]
\begin{adjustbox}{width=1\textwidth}
\begin{tabular}{@{}l|rrrrrrrrrrrr@{}}
\toprule
\textbf{Features\textbackslash{}Layers} & 1 & 2 & 3 & 4 & 5 & 6 & 7 & 8 & 9 & 10 & 11 & 12 \\ \midrule
filled\_pause\_rate & 0.0000064 & 0.0000087 & 0.0000047 & 0.0000042 & 0.0000092 & 0.0000036 & 0.0000056 & 0.0000265 & 0.0000234 & 0.0000037 & \textbf{0.0000034} & 0.0000037 \\
general\_silence & 0.010336 & 0.009149 & 0.009781 & 0.009449 & 0.010214 & 0.009557 & 0.010462 & 0.01013 & 0.009626 & 0.0107 & \textbf{0.00914} & 0.010858 \\
mean\_silence & 0.008367 & 0.00818 & 0.008529 & \textbf{0.007218} & 0.007962 & 0.008569 & 0.007368 & 0.008397 & 0.00864 & 0.008453 & 0.009267 & 0.00882 \\
silence\_abs\_deviation & 0.008834 & 0.008651 & 0.008101 & 0.008431 & 0.008082 & 0.008105 & 0.008496 & 0.007868 & \textbf{0.007671} & 0.008466 & 0.009121 & 0.010871 \\
SilenceRate1 & 0.010441 & 0.009092 & 0.010433 & 0.009222 & 0.009357 & 0.008964 & 0.009186 & 0.009738 & 0.010078 & 0.00949 & \textbf{0.008925} & 0.010444 \\
SilenceRate2 & 0.019169 & \textbf{0.017716} & 0.01913 & 0.018713 & 0.018604 & 0.017923 & 0.018491 & 0.019281 & 0.018252 & 0.018755 & 0.017999 & 0.022653 \\
speaking\_rate & 0.009495 & 0.008707 & 0.00915 & 0.009289 & 0.009674 & 0.008421 & 0.007933 & 0.008018 & \textbf{0.007882} & 0.009373 & 0.010031 & 0.009356 \\
articulation\_rate & 0.014608 & 0.012277 & 0.011997 & 0.012936 & 0.012124 & \textbf{0.011002} & 0.011027 & 0.011786 & 0.01273 & 0.011808 & 0.012298 & 0.012386 \\
longpfreq & 0.006085 & 0.00566 & 0.00515 & 0.00531 & \textbf{0.004731} & 0.005081 & 0.005215 & 0.005338 & 0.005203 & 0.004949 & 0.005599 & 0.005909 \\
average\_syllables\_in\_words & 0.040541 & 0.039341 & 0.038896 & 0.040124 & 0.034919 & 0.032039 & \textbf{0.026923} & 0.0319 & 0.030701 & 0.034601 & 0.046442 & 0.040721 \\
wordsyll2 & 0.02971 & 0.029581 & 0.029094 & 0.027793 & 0.027336 & 0.026798 & \textbf{0.02158} & 0.023803 & 0.025616 & 0.02566 & 0.030421 & 0.028458 \\
repetition\_freq & 0.025743 & 0.026271 & 0.026023 & 0.026202 & 0.026274 & \textbf{0.025517} & 0.025752 & 0.027057 & 0.026717 & 0.025908 & 0.026287 & 0.026288\\ \bottomrule
\end{tabular}
\end{adjustbox}
\vspace{1 mm}
\caption{\label{nFluency_W}  \small Results (MSE) for fluency features on {\wv} for non-native read speech corpus (L2 Arctic)}

\end{table*}

\begin{table*}[]
\begin{adjustbox}{width=1\textwidth}
\begin{tabular}{@{}l|rrrrrrrrrrrr@{}}
\toprule
\textbf{Features\textbackslash{}Layers} & 1 & 2 & 3 & 4 & 5 & 6 & 7 & 8 & 9 & 10 & 11 & 12 \\ \midrule
StressDistanceSyllMean & 0.010858 & 0.010986 & 0.011084 & 0.010706 & \textbf{0.010447} & 0.010925 & 0.01058 & 0.010629 & 0.010961 & 0.010965 & 0.010822 & 0.010862 \\
StressDistanceMean & 0.01445 & 0.01437 & 0.014744 & 0.014552 & 0.014447 & 0.014385 & \textbf{0.013805} & 0.014652 & 0.014157 & 0.014463 & 0.014664 & 0.014562 \\
vowelPercentage & 0.006582 & 0.006036 & 0.005799 & 0.005019 & 0.004906 & 0.005743 & 0.005376 & 0.005848 & \textbf{0.004815} & 0.005328 & 0.006171 & 0.006102 \\
consonantPercentage & 0.010776 & 0.009302 & 0.009351 & 0.011724 & 0.007659 & \textbf{0.006691} & 0.008678 & 0.008015 & 0.008811 & 0.008519 & 0.009191 & 0.009436 \\
vowelDurationSD & 0.004968 & 0.005062 & 0.004467 & 0.004498 & 0.004241 & \textbf{0.004157} & 0.00432 & 0.004319 & 0.004306 & 0.004624 & 0.004953 & 0.004792 \\
consonantDurationSD & 0.008937 & 0.008598 & 0.008309 & 0.008176 & 0.008196 & \textbf{0.007876} & 0.008254 & 0.008291 & 0.008224 & 0.00824 & 0.009342 & 0.008896 \\
syllableDurationSD & 0.018064 & 0.017007 & 0.016705 & 0.015943 & 0.016991 & 0.015742 & \textbf{0.015609} & 0.016092 & 0.016 & 0.016202 & 0.017674 & 0.019068 \\
vowelSDNorm & 0.00729 & 0.007269 & 0.007573 & 0.007196 & 0.007126 & 0.007205 & 0.00711 & 0.007452 & \textbf{0.007106} & 0.007388 & 0.007281 & 0.007137 \\
consonantSDNorm & 0.011117 & 0.010125 & 0.010393 & 0.010053 & 0.010329 & 0.01008 & 0.010348 & 0.010509 & \textbf{0.010004} & 0.011252 & 0.011404 & 0.011133 \\
syllableSDNorm & 0.015935 & 0.016013 & 0.015636 & 0.015655 & 0.017318 & 0.01532 & \textbf{0.014836} & 0.016397 & 0.017799 & 0.015735 & 0.01658 & 0.017679 \\
vowelPVINorm & 0.006011 & 0.006213 & 0.006066 & 0.006023 & 0.005908 & 0.005946 & \textbf{0.0058} & 0.005853 & 0.005862 & 0.00597 & 0.00606 & 0.005956 \\
consonantPVINorm & \textbf{0.009807} & 0.009825 & 0.010233 & 0.010364 & 0.010156 & 0.01187 & 0.009982 & 0.010278 & 0.01084 & 0.011683 & 0.010663 & 0.010516 \\
syllablePVINorm & 0.015131 & 0.014699 & 0.014601 & 0.014249 & 0.015095 & 0.015608 & \textbf{0.01386} & 0.015136 & 0.014716 & 0.01435 & 0.015585 & 0.014903 \\ \bottomrule
\end{tabular}
\end{adjustbox}
\vspace{1 mm}
\caption{\label{nPron_W} \small  Results (MSE) for pronunciation features on {\wv} for non-native read speech corpus (L2 Arctic)}

\end{table*}

\begin{table*}[]
\begin{adjustbox}{width=1\textwidth}
\begin{tabular}{@{}l|rrrrrrrrrrrr@{}}
\toprule
\textbf{Features\textbackslash{}Layers} & 1        & 2        & 3                 & 4        & 5        & 6                 & 7                 & 8                 & 9                 & 10       & 11       & 12       \\
Unique Word count                       & 0.016328 & 0.015269 & 0.01443           & 0.014206 & 0.016187 & 0.016495          & \textbf{0.013196} & 0.01333           & 0.014827          & 0.015252 & 0.016198 & 0.018124 \\
Word Complexity                         & 0.02358  & 0.025074 & 0.026571          & 0.023607 & 0.023869 & \textbf{0.022757} & 0.024556          & 0.023255          & 0.024087          & 0.02473  & 0.025855 & 0.026026 \\
Total adjectives                        & 0.055104 & 0.055733 & 0.055438          & 0.0582   & 0.053788 & 0.053288          & 0.051659          & \textbf{0.051016} & 0.055142          & 0.05502  & 0.058229 & 0.058659 \\
Total adverbs                           & 0.032355 & 0.031032 & \textbf{0.030508} & 0.031414 & 0.034148 & 0.03062           & 0.033625          & 0.031246          & 0.032231          & 0.0325   & 0.030786 & 0.030738 \\
Total nouns                             & 0.031529 & 0.032214 & 0.034053          & 0.032932 & 0.032949 & 0.029525          & \textbf{0.028352} & 0.029281          & 0.029507          & 0.030491 & 0.032188 & 0.033265 \\
Total verbs                             & 0.038064 & 0.037304 & 0.041601          & 0.037693 & 0.03675  & 0.040773          & \textbf{0.035506} & 0.037434          & 0.045298          & 0.040241 & 0.037563 & 0.037659 \\
Total pronoun                           & 0.019591 & 0.019544 & 0.01956           & 0.019596 & 0.019562 & \textbf{0.019445} & 0.019699          & 0.019446          & 0.019506          & 0.019491 & 0.019485 & 0.019494 \\
Total conjunction                       & 0.042578 & 0.042498 & 0.042017          & 0.041322 & 0.042358 & 0.042192          & 0.041659          & 0.041382          & \textbf{0.041127} & 0.041878 & 0.042459 & 0.042781 \\
Total determiners                       & 0.01978  & 0.019613 & 0.019911          & 0.01972  & 0.019727 & 0.019582          & 0.019646          & \textbf{0.019535} & 0.01958           & 0.019619 & 0.01961  & 0.019776 \\
No. of subj                             & 0.067824 & 0.065793 & 0.065282          & 0.066421 & 0.064326 & 0.065557          & \textbf{0.061022} & 0.061545          & 0.064313          & 0.065425 & 0.067203 & 0.068783 \\
No. of obj                              & 0.04038  & 0.040672 & 0.040469          & 0.041629 & 0.040958 & 0.039576          & 0.037172          & \textbf{0.036589} & 0.038272          & 0.040714 & 0.040737 & 0.042256 \\
Tree depth                              & 0.021626 & 0.021995 & 0.02262           & 0.022675 & 0.022558 & 0.021697          & \textbf{0.020467} & 0.021307          & 0.024375          & 0.023414 & 0.022422 & 0.023464 \\
 \bottomrule
\end{tabular}
\end{adjustbox}
\vspace{1 mm}
\caption{\label{nVocab_W} \small  Results (MSE) for text features on {\wv} for non-native read speech corpus (L2 Arctic)}
\end{table*}

\begin{table*}[]
\begin{adjustbox}{width=1\textwidth}
\begin{tabular}{@{}l|rrrrrrrrrrrr@{}}
\toprule
\textbf{Features\textbackslash{}Layers} & 1 & 2 & 3 & 4 & 5 & 6 & 7 & 8 & 9 & 10 & 11 & 12 \\ \midrule
total\_duration & 0.011324 & \textbf{0.003763} & 0.007239 & 0.005672 & 0.005221 & 0.007031 & 0.007505 & 0.007055 & 0.007161 & 0.007567 & 0.010025 & 0.012677 \\
stdev\_energy & 0.015313 & 0.015339 & 0.016494 & 0.014493 & 0.01403 & 0.015362 & 0.013972 & 0.017967 & 0.014863 & 0.013474 & \textbf{0.011171} & 0.012512 \\
mean\_pitch & 0.016319 & 0.017864 & 0.013842 & 0.014276 & 0.011206 & 0.0148 & 0.017934 & 0.011364 & 0.008506 & 0.003965 & \textbf{0.003368} & 0.003793 \\
voiced\_to\_unvoiced\_ratio & 0.002841 & 0.002697 & 0.002844 & 0.002684 & 0.002754 & 0.002926 & 0.00292 & 0.002638 & 0.002202 & 0.001792 & \textbf{0.00161} & 0.002214 \\
zero\_crossing\_rate & 0.011983 & 0.013096 & 0.011805 & 0.012052 & 0.011352 & 0.013772 & 0.017485 & 0.012387 & 0.007931 & 0.007328 & \textbf{0.004908} & 0.005565 \\
energy\_entropy & 0.014261 & 0.013957 & \textbf{0.011386} & 0.011758 & 0.013304 & 0.012495 & 0.015632 & 0.014125 & 0.016826 & 0.012122 & 0.013926 & 0.015025 \\
spectral\_centroid & 0.000003 & 0.000003 & 0.000003 & 0.000003 & 0.000004 & \textbf{0.000003} & 0.000005 & 0.000003 & 0.000003 & 0.000003 & 0.000003 & 0.000005 \\
localJitter & 0.009555 & 0.009915 & 0.008388 & 0.009049 & 0.00897 & 0.011487 & 0.011618 & 0.007706 & 0.007508 & 0.007772 & 0.007588 & \textbf{0.007487} \\
localShimmer & 0.007763 & 0.007366 & 0.0061 & 0.006087 & 0.006472 & 0.006134 & 0.007093 & 0.00579 & 0.006633 & 0.005442 & 0.006895 & \textbf{0.004808}\\ \bottomrule
\end{tabular}
\end{adjustbox}
\vspace{1 mm}
\caption{\label{nAudio_M} \small  Results (MSE) for audio features on {\mj} for non-native read speech corpus (L2 Arctic)}

\end{table*}

\begin{table*}[]
\begin{adjustbox}{width=1\textwidth}
\begin{tabular}{@{}l|rrrrrrrrrrrr@{}}
\toprule
\textbf{Features\textbackslash{}Layers} & 1 & 2 & 3 & 4 & 5 & 6 & 7 & 8 & 9 & 10 & 11 & 12 \\ \midrule
filled\_pause\_rate & 0.0000015 & 0.0000054 & 0.0000017 & 0.0000085 & 0.0000039 & 0.0000132 & \textbf{0.0000001} & 0.0000134 & 0.0000775 & 0.0000016 & 0.0000036 & 0.0000513 \\
general\_silence & 0.013079 & 0.011675 & 0.011336 & 0.010921 & 0.010777 & 0.010924 & 0.012546 & 0.011623 & 0.01091 & \textbf{0.009547} & 0.011005 & 0.012333 \\
mean\_silence & 0.009038 & 0.008802 & 0.00913 & 0.009513 & 0.010127 & 0.011607 & 0.010981 & 0.008231 & 0.007322 & 0.008047 & \textbf{0.007277} & 0.007887 \\
silence\_abs\_deviation & 0.008988 & 0.007935 & 0.009433 & 0.009917 & 0.008996 & 0.009184 & 0.009437 & 0.008632 & \textbf{0.007513} & 0.00863 & 0.008548 & 0.009037 \\
SilenceRate1 & 0.011366 & 0.0108 & 0.011012 & 0.009303 & 0.009495 & 0.010584 & 0.009934 & 0.009442 & 0.009309 & 0.0089 & \textbf{0.008661} & 0.009029 \\
SilenceRate2 & 0.020763 & 0.019909 & 0.019526 & 0.018436 & 0.021103 & 0.018528 & 0.019417 & 0.018365 & 0.017279 & 0.017109 & \textbf{0.01662} & 0.018189 \\
speaking\_rate & 0.010575 & 0.010034 & 0.009617 & 0.009627 & 0.009733 & 0.010429 & 0.013854 & 0.01039 & 0.010063 & 0.011578 & 0.010152 & \textbf{0.009589} \\
articulation\_rate & 0.015664 & 0.014572 & 0.015364 & 0.014105 & 0.013773 & 0.015194 & 0.017737 & 0.016386 & 0.015303 & 0.01375 & \textbf{0.013596} & 0.015014 \\
longpfreq & 0.007634 & 0.006734 & 0.009045 & 0.006316 & 0.005609 & 0.006313 & 0.006256 & 0.005071 & 0.004951 & 0.004762 & \textbf{0.004741} & 0.004901 \\
average\_syllables\_in\_words & 0.047811 & 0.053749 & \textbf{0.041714} & 0.043455 & 0.044667 & 0.047274 & 0.043626 & 0.044136 & 0.043008 & 0.04317 & 0.043801 & 0.045771 \\
wordsyll2 & 0.037088 & 0.03632 & 0.035302 & \textbf{0.034771} & 0.037232 & 0.038009 & 0.036853 & 0.037443 & 0.03709 & 0.035578 & 0.036575 & 0.035855 \\
repetition\_freq & 0.02641 & 0.026258 & 0.026411 & 0.026243 & \textbf{0.026179} & 0.026339 & 0.02649 & 0.026482 & 0.026486 & 0.026519 & 0.026192 & 0.026305 \\ \bottomrule
\end{tabular}
\end{adjustbox}
\vspace{1 mm}
\caption{\label{nFluency_M} \small  Results (MSE) for fluency features on {\mj} for non-native read speech corpus (L2 Arctic)}

\end{table*}

\begin{table*}[]
\begin{adjustbox}{width=1\textwidth}
\begin{tabular}{@{}l|rrrrrrrrrrrr@{}}
\toprule
\textbf{Features\textbackslash{}Layers} & 1 & 2 & 3 & 4 & 5 & 6 & 7 & 8 & 9 & 10 & 11 & 12 \\  \midrule
StressDistanceSyllMean & 0.0110454 & 0.011099 & 0.0113248 & 0.0108703 & 0.0109954 & \textbf{0.0108252} & 0.0109959 & 0.0110083 & 0.0110286 & 0.0111566 & 0.0108324 & 0.0112061 \\
StressDistanceMean & 0.015089 & 0.015405 & 0.014999 & 0.015803 & \textbf{0.014757} & 0.015073 & 0.015133 & 0.015133 & 0.015081 & 0.014896 & 0.015153 & 0.015847 \\
vowelPercentage & 0.007385 & 0.007232 & 0.007148 & 0.006743 & 0.009473 & 0.007247 & 0.007359 & 0.006418 & 0.006181 & 0.005328 & \textbf{0.005268} & 0.005668 \\
consonantPercentage & 0.011632 & 0.012139 & 0.010961 & 0.011663 & 0.010474 & 0.01305 & 0.014139 & 0.011154 & 0.008865 & \textbf{0.008217} & 0.008636 & 0.011022 \\
vowelDurationSD & 0.005576 & 0.005649 & 0.005741 & 0.005688 & 0.005604 & 0.005607 & 0.00567 & 0.005818 & 0.005529 & 0.005221 & \textbf{0.004916} & 0.005037 \\
consonantDurationSD & 0.009749 & 0.009875 & 0.009595 & 0.009228 & 0.009425 & 0.009572 & 0.009824 & 0.009433 & 0.009011 & 0.008617 & \textbf{0.008287} & 0.008859 \\
syllableDurationSD & 0.02133 & 0.021847 & 0.022648 & 0.019743 & 0.020138 & 0.020474 & 0.021925 & 0.02083 & 0.020653 & 0.018907 & \textbf{0.017768} & 0.018129 \\
vowelSDNorm & 0.007599 & 0.007843 & 0.008135 & 0.007658 & 0.007602 & 0.007557 & 0.007632 & 0.007809 & 0.007583 & \textbf{0.007363} & 0.007512 & 0.007515 \\
consonantSDNorm & 0.012147 & 0.012031 & 0.011663 & 0.012369 & 0.01163 & 0.011643 & 0.011955 & 0.012129 & 0.011969 & \textbf{0.010975} & 0.011351 & 0.01106 \\
syllableSDNorm & 0.018041 & 0.016371 & 0.018828 & 0.019273 & \textbf{0.016217} & 0.016602 & 0.016867 & 0.016381 & 0.016579 & 0.017109 & 0.017468 & 0.016904 \\
vowelPVINorm & 0.006216 & 0.006449 & 0.007166 & 0.006115 & 0.006628 & 0.0061 & 0.006222 & 0.006118 & 0.006216 & 0.006102 & \textbf{0.006067} & 0.006612 \\
consonantPVINorm & 0.011656 & 0.010482 & 0.01176 & 0.011887 & 0.010422 & 0.011001 & 0.010469 & 0.010507 & 0.0106 & 0.011769 & \textbf{0.010104} & 0.010276 \\
syllablePVINorm & 0.01476 & 0.014777 & 0.015596 & \textbf{0.014607} & 0.015637 & 0.014877 & 0.014862 & 0.014626 & 0.01491 & 0.014709 & 0.015419 & 0.014619 \\ \bottomrule
\end{tabular}
\end{adjustbox}
\vspace{1 mm}
\caption{\label{nPron_M} \small  Results (MSE) for pronunciation features on {\mj} for non-native read speech corpus (L2 Arctic)}

\end{table*}

\begin{table*}[]
\begin{adjustbox}{width=1\textwidth}
\begin{tabular}{@{}l|rrrrrrrrrrrr@{}}
\toprule
\textbf{Features\textbackslash{}Layers} & 1                 & 2                 & 3        & 4                 & 5                 & 6        & 7        & 8                 & 9        & 10                & 11                & 12       \\
Unique Word count                       & 0.019898          & \textbf{0.017283} & 0.017832 & 0.024237          & 0.018815          & 0.019755 & 0.022221 & 0.022647          & 0.020655 & 0.024299          & 0.026161          & 0.028878 \\
Word Complexity                         & 0.025144          & 0.02611           & 0.024706 & 0.027259          & 0.024931          & 0.024704 & 0.026681 & 0.024872          & 0.024702 & 0.025301          & \textbf{0.024337} & 0.025369 \\
Total adjectives                        & 0.063184          & 0.061925          & 0.060797 & 0.060141          & \textbf{0.059878} & 0.060538 & 0.06247  & 0.063014          & 0.063618 & 0.061781          & 0.062812          & 0.063825 \\
Total adverbs                           & 0.031172          & 0.030957          & 0.030912 & 0.031867          & 0.030752          & 0.031637 & 0.03176  & \textbf{0.030206} & 0.030851 & 0.031676          & 0.03124           & 0.031381 \\
Total nouns                             & 0.034793          & 0.041242          & 0.035093 & \textbf{0.034609} & 0.034843          & 0.039378 & 0.03622  & 0.039652          & 0.035273 & 0.034804          & 0.036142          & 0.039605 \\
Total verbs                             & \textbf{0.038152} & 0.038439          & 0.040985 & 0.039095          & 0.041912          & 0.042645 & 0.039212 & 0.038552          & 0.039155 & 0.038542          & 0.043597          & 0.039812 \\
Total pronoun                           & 0.019712          & 0.019852          & 0.019967 & 0.019653          & 0.019711          & 0.019769 & 0.019644 & \textbf{0.019642} & 0.019674 & 0.019667          & 0.019718          & 0.019797 \\
Total conjunction                       & 0.042722          & \textbf{0.041393} & 0.044539 & 0.042758          & 0.042504          & 0.042034 & 0.042042 & 0.041905          & 0.042218 & 0.042248          & 0.042221          & 0.042478 \\
Total determiners                       & 0.019733          & 0.019827          & 0.019754 & 0.019708          & 0.019816          & 0.019727 & 0.019809 & 0.019789          & 0.019792 & \textbf{0.019536} & 0.019631          & 0.019608 \\
No. of subj                             & 0.068228          & 0.067126          & 0.065661 & 0.065894          & 0.066666          & 0.065434 & 0.065386 & 0.066936          & 0.066377 & \textbf{0.065255} & 0.066918          & 0.066453 \\
No. of obj                              & 0.041227          & 0.041092          & 0.04096  & 0.041113          & 0.042383          & 0.041107 & 0.041013 & 0.041913          & 0.040938 & \textbf{0.040895} & 0.041729          & 0.041005 \\
Tree depth                              & 0.025486          & \textbf{0.023126} & 0.023168 & 0.024691          & 0.023127          & 0.023284 & 0.024154 & 0.024006          & 0.02503  & 0.023679          & 0.023465          & 0.02412 \\ \bottomrule
\end{tabular}
\end{adjustbox}
\vspace{1 mm}
\caption{\label{nVocab_M} \small  Results (MSE) for text features on {\mj} for non-native read speech corpus (L2 Arctic)}

\end{table*}

\begin{table*}[]
\begin{adjustbox}{width=1\textwidth}
\begin{tabular}{@{}l|rrrrrrrrrrrr@{}}
\toprule
\textbf{Features\textbackslash{}Layers} & 1 & 2 & 3 & 4 & 5 & 6 & 7 & 8 & 9 & 10 & 11 & 12 \\ \midrule
total\_duration & 0.0048489 & 0.0038639 & 0.0042871 & \textbf{0.0037838} & 0.0043445 & 0.0044276 & 0.0038112 & 0.0051095 & 0.0048823 & 0.0057632 & 0.0099903 & 0.0075039 \\
stdev\_energy & 0.017303 & \textbf{0.016815} & 0.017969 & 0.019661 & 0.020539 & 0.020834 & 0.0208 & 0.020277 & 0.019462 & 0.019713 & 0.021847 & 0.019478 \\
mean\_pitch & 0.01006 & \textbf{0.006586} & 0.008036 & 0.008113 & 0.008285 & 0.008348 & 0.010017 & 0.01013 & 0.009512 & 0.009188 & 0.013511 & 0.011926 \\
voiced\_to\_unvoiced\_ratio & 0.006834 & 0.007377 & 0.005871 & 0.005814 & 0.007002 & 0.006451 & 0.007266 & \textbf{0.005397} & 0.006472 & 0.007256 & 0.008058 & 0.007152 \\
zero\_crossing\_rate & 0.013572 & 0.01401 & \textbf{0.013133} & 0.013668 & 0.013948 & 0.015045 & 0.014912 & 0.015759 & 0.013684 & 0.013544 & 0.019732 & 0.014926 \\
energy\_entropy & 0.012869 & 0.014251 & 0.013436 & 0.013081 & 0.013873 & \textbf{0.012395} & 0.015116 & 0.013598 & 0.016164 & 0.019665 & 0.015759 & 0.017393 \\
spectral\_centroid & \textbf{0.004404} & 0.004427 & 0.004423 & 0.004409 & 0.00442 & 0.004409 & 0.004418 & 0.00442 & 0.004421 & 0.004421 & 0.004424 & 0.004407 \\
localJitter & \textbf{0.01373} & 0.01405 & 0.013894 & 0.014942 & 0.014555 & 0.014971 & 0.014774 & 0.015739 & 0.014834 & 0.014159 & 0.015629 & 0.014187 \\
localShimmer & 0.012059 & 0.009315 & 0.010288 & 0.009537 & \textbf{0.009271} & 0.010519 & 0.010244 & 0.010275 & 0.009516 & 0.009711 & 0.011389 & 0.010449\\ \bottomrule
\end{tabular}
\end{adjustbox}
\vspace{1 mm}
\caption{\label{nsAudio_W} \small  Results (MSE) for audio features on {\wv} for native spontaneous speech corpus (Mozilla Common Voice)}

\end{table*}

\begin{table*}[]
\begin{adjustbox}{width=1\textwidth}
\begin{tabular}{@{}l|rrrrrrrrrrrr@{}}
\toprule
\textbf{Features\textbackslash{}Layers} & 1 & 2 & 3 & 4 & 5 & 6 & 7 & 8 & 9 & 10 & 11 & 12 \\ \midrule
Unique Word count & 0.018459 & 0.020039 & 0.020034 & 0.017761 & 0.022014 & 0.018712 & 0.018018 & 0.020214 & 0.019047 & 0.018931 & 0.021376 & \textbf{0.017354} \\
Word Complexity & 0.017605 & 0.016675 & 0.016306 & 0.019499 & 0.017307 & 0.019961 & 0.017373 & 0.020203 & \textbf{0.015218} & 0.01758 & 0.016081 & 0.015885\\ 
Total adjectives                        & 0.035461          & 0.033096          & 0.031345          & 0.032745          & 0.031163          & 0.033431 & 0.033394 & 0.031849 & 0.033459          & \textbf{0.03104} & 0.033272          & 0.034714 \\
Total adverbs                           & 0.034797          & 0.03482           & 0.034696          & 0.034693          & 0.034678          & 0.034757 & 0.034598 & 0.034654 & \textbf{0.034556} & 0.034625         & 0.034889          & 0.034863 \\
Total nouns                             & 0.020902          & 0.01946           & 0.02546           & 0.018685          & \textbf{0.018604} & 0.018838 & 0.021946 & 0.021027 & 0.021565          & 0.026668         & 0.02171           & 0.022908 \\
Total verbs                             & \textbf{0.013959} & 0.014957          & 0.014028          & 0.01421           & 0.014333          & 0.014363 & 0.014145 & 0.014215 & 0.014078          & 0.014181         & 0.014571          & 0.014432 \\
Total pronoun                           & 0.004981          & 0.004971          & 0.004946          & \textbf{0.004937} & 0.004983          & 0.00496  & 0.00498  & 0.004989 & 0.004983          & 0.004987         & 0.004985          & 0.004985 \\
Total conjunction                       & 0.018495          & 0.018835          & \textbf{0.018418} & 0.018694          & 0.018758          & 0.018616 & 0.018439 & 0.018429 & 0.018532          & 0.018459         & 0.018465          & 0.018456 \\
Total determiners                       & 0.001366          & 0.001276          & 0.001313          & 0.001263          & 0.001305          & 0.001283 & 0.001324 & 0.001254 & 0.001257          & 0.001276         & \textbf{0.001246} & 0.001384 \\
No. of subj                             & 0.023393          & 0.023389          & 0.02298           & 0.023171          & 0.024021          & 0.02304  & 0.023167 & 0.022995 & 0.02307           & 0.023339         & \textbf{0.022829} & 0.023223 \\
No. of obj                              & 0.032886          & 0.033575          & 0.032966          & 0.033214          & 0.033236          & 0.033246 & 0.034028 & 0.033074 & 0.033186          & \textbf{0.0328}  & 0.03292           & 0.033067 \\
Tree depth                              & 0.020084          & 0.021207          & 0.021134          & \textbf{0.019598} & 0.020404          & 0.020362 & 0.020963 & 0.021873 & 0.020284          & 0.019893         & 0.020609          & 0.024622 \\
\bottomrule
\end{tabular}
\end{adjustbox}
\vspace{1 mm}
\caption{\label{nsVocab_W} \small  Results (MSE) for text features on {\wv} for native spontaneous speech corpus (Mozilla Common Voice)}
\end{table*}

\begin{table*}[]
\begin{adjustbox}{width=1\textwidth}
\begin{tabular}{@{}l|rrrrrrrrrrrr@{}}
\toprule
\textbf{Features\textbackslash{}Layers} & 1 & 2 & 3 & 4 & 5 & 6 & 7 & 8 & 9 & 10 & 11 & 12 \\ \midrule
total\_duration & 0.019696 & 0.009559 & \textbf{0.009083} & 0.010356 & 0.009995 & 0.010414 & 0.014136 & 0.012907 & 0.017391 & 0.017447 & 0.02315 & 0.03311 \\
stdev\_energy & 0.023013 & 0.023561 & 0.023576 & 0.023124 & 0.023222 & 0.022773 & 0.023324 & 0.023156 & 0.022229 & \textbf{0.020345} & 0.020714 & 0.022506 \\
mean\_pitch & 0.031958 & 0.032023 & 0.03432 & 0.031613 & 0.035276 & 0.033878 & 0.038073 & 0.030162 & 0.026041 & 0.017525 & 0.015405 & \textbf{0.013203} \\
voiced\_to\_unvoiced\_ratio & 0.011325 & 0.012245 & 0.011508 & 0.01051 & 0.010533 & 0.01126 & 0.011826 & 0.009867 & 0.008246 & 0.00643 & \textbf{0.005966} & 0.007566 \\
zero\_crossing\_rate & 0.025145 & 0.023447 & 0.022659 & 0.023914 & 0.023128 & 0.025615 & 0.024162 & 0.021813 & 0.019922 & 0.019591 & \textbf{0.017914} & 0.01907 \\
energy\_entropy & 0.02261 & 0.018648 & 0.019247 & 0.022381 & 0.024713 & 0.0281 & 0.022333 & 0.023445 & 0.020404 & \textbf{0.015934} & 0.024606 & 0.021252 \\
spectral\_centroid & 0.00442 & 0.00442 & 0.004416 & 0.004421 & 0.004414 & 0.004421 & 0.00442 & 0.004423 & 0.004421 & 0.00442 & \textbf{0.004399} & 0.004422 \\
localJitter & 0.018309 & 0.018079 & 0.018014 & 0.018241 & 0.017157 & 0.018334 & 0.01813 & 0.017893 & 0.016641 & 0.015908 & \textbf{0.014427} & 0.015825 \\
localShimmer & 0.01804 & 0.016669 & 0.016321 & 0.016724 & 0.017018 & 0.016635 & 0.016631 & 0.016908 & 0.015111 & 0.013504 & \textbf{0.012347} & 0.013062\\ \bottomrule
\end{tabular}
\end{adjustbox}
\vspace{1 mm}
\caption{\label{nsAudio_M}  \small Results (MSE) for audio features on {\mj} for native spontaneous speech corpus (Mozilla Common Voice)}

\end{table*}

\begin{table*}[]
\begin{adjustbox}{width=1\textwidth}
\begin{tabular}{@{}l|rrrrrrrrrrrr@{}}
\toprule
\textbf{Features\textbackslash{}Layers} & 1 & 2 & 3 & 4 & 5 & 6 & 7 & 8 & 9 & 10 & 11 & 12 \\ \midrule
Unique Word count & 0.018624 & 0.018947 & 0.018667 & 0.018967 & 0.019152 & 0.018748 & \textbf{0.018482} & 0.020656 & 0.019855 & 0.020164 & 0.018702 & 0.018696 \\
Word Complexity & \textbf{0.015312} & 0.015486 & 0.017034 & 0.019543 & 0.018866 & 0.017361 & 0.018065 & 0.017269 & 0.015957 & 0.015654 & 0.017122 & 0.018025\\
Total adjectives                        & 0.033444 & 0.028188 & 0.027305 & 0.027058          & 0.026591 & 0.026863 & 0.025666          & \textbf{0.025522} & 0.025805 & 0.027863 & 0.032032          & 0.028127          \\
Total adverbs                           & 0.034442 & 0.035042 & 0.034444 & 0.034427          & 0.03449  & 0.033877 & 0.03448           & \textbf{0.033116} & 0.034165 & 0.034304 & 0.034875          & 0.034641          \\
Total nouns                             & 0.018199 & 0.015158 & 0.015353 & 0.013575          & 0.013938 & 0.016498 & \textbf{0.012826} & 0.015598          & 0.014673 & 0.014727 & 0.018801          & 0.017372          \\
Total verbs                             & 0.013491 & 0.013059 & 0.01343  & 0.013238          & 0.014039 & 0.013877 & 0.013065          & \textbf{0.012314} & 0.014256 & 0.013209 & 0.014094          & 0.013936          \\
Total pronoun                           & 0.004965 & 0.004974 & 0.004984 & 0.004969          & 0.004968 & 0.00497  & 0.004972          & 0.004964          & 0.005011 & 0.004995 & 0.004979          & \textbf{0.004958} \\
Total conjunction                       & 0.018383 & 0.018359 & 0.018387 & \textbf{0.018346} & 0.018373 & 0.018376 & 0.018407          & 0.018403          & 0.018666 & 0.018489 & 0.018441          & 0.018413          \\
Total determiners                       & 0.00125  & 0.001309 & 0.001427 & 0.001296          & 0.001268 & 0.001323 & 0.001275          & 0.001272          & 0.001254 & 0.001254 & \textbf{0.001244} & 0.001295          \\
No. of subj                             & 0.02277  & 0.022932 & 0.023061 & 0.022751          & 0.022702 & 0.022716 & \textbf{0.022602} & 0.022962          & 0.023014 & 0.022769 & 0.023416          & 0.022838          \\
No. of obj                              & 0.032844 & 0.032739 & 0.032577 & 0.032527          & 0.033305 & 0.033533 & 0.032881          & 0.032321          & 0.032717 & 0.033183 & 0.032922          & \textbf{0.031866} \\
Tree depth                              & 0.018059 & 0.018037 & 0.018863 & 0.018194          & 0.01816  & 0.017753 & 0.017491          & \textbf{0.016916} & 0.018079 & 0.018034 & 0.019362          & 0.018806 \\
\bottomrule
\end{tabular}
\end{adjustbox}
\vspace{1 mm}
\caption{\label{nsVocab_M} \small Results (MSE) for text features on {\mj} for native spontaneous speech corpus (Mozilla Common Voice)}
\end{table*}

\end{document}